\documentclass[letterpaper]{article} 
\usepackage{aaai24}  
\usepackage{times}  
\usepackage{helvet}  
\usepackage{courier}  
\usepackage[hyphens]{url}  
\usepackage{graphicx} 
\usepackage{natbib}  
\usepackage{caption} 
\usepackage{algorithm}
\usepackage{algorithmic}

\usepackage{algorithm}
\usepackage{algorithmic}
\usepackage{amsmath}
\usepackage{amssymb}
\usepackage{amsthm}
\usepackage{booktabs,subcaption,amsfonts,dcolumn}
\newtheorem{theorem}{Theorem}
\usepackage{times}
\usepackage{epsfig}
\usepackage{graphicx}
\usepackage{subcaption}

\usepackage{listings}
\usepackage{comment}
\usepackage{multirow}
\usepackage{microtype}  
\usepackage{xcolor}

\DeclareMathOperator*{\argminA}{arg\,min}

\usepackage{newfloat}
\usepackage{listings}
\DeclareCaptionStyle{ruled}{labelfont=normalfont,labelsep=colon,strut=off} 
\floatstyle{ruled}
\newfloat{listing}{tb}{lst}{}
\floatname{listing}{Listing}
\title{Foreseeing Reconstruction Quality of \\ Gradient Inversion: An Optimization Perspective}
\author{
    HyeongGwon Hong\textsuperscript{\rm 1},
    Yooshin Cho\textsuperscript{\rm 2}, 
    Hanbyel Cho\textsuperscript{\rm 2}, 
    Jaesung Ahn\textsuperscript{\rm 1}, 
    Junmo Kim\textsuperscript{\rm 2}
}
\affiliations{
    \textsuperscript{\rm 1}Kim Jaechul Graduate School of AI, KAIST, Seoul, South Korea\\
    \textsuperscript{\rm 2}School of Electrical Engineering, KAIST, Daejeon, South Korea\\

    \{honggudrnjs, choys95, tlrl4658, jaesung02, junmo.kim\}@kaist.ac.kr
}

\begin{document}

\maketitle

\begin{figure*}
    \centering
    \begin{subfigure}[b]{0.2\textwidth}
        \includegraphics[width=\linewidth]{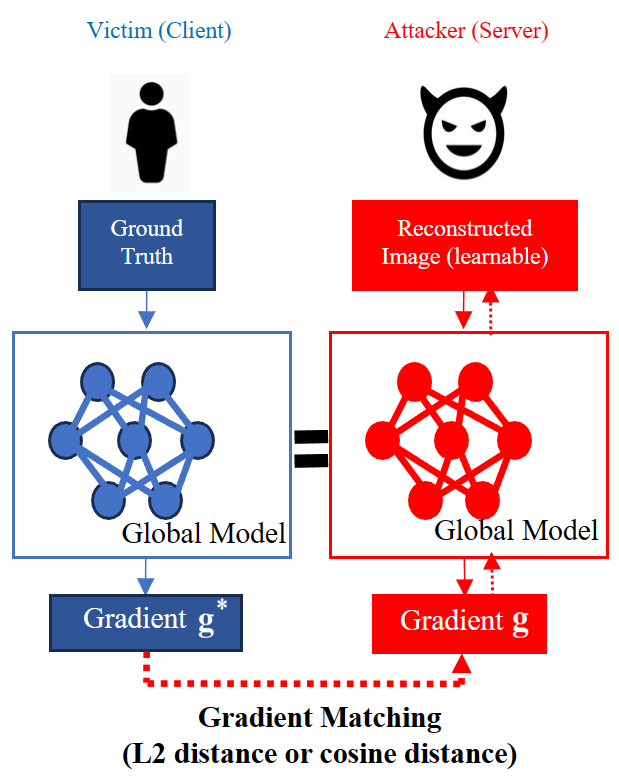}
        \caption{An overview of gradient inversion attack}
        \label{fig:c1}
    \end{subfigure}
    \begin{subfigure}[b]{0.37\textwidth}
        \includegraphics[width=\linewidth]{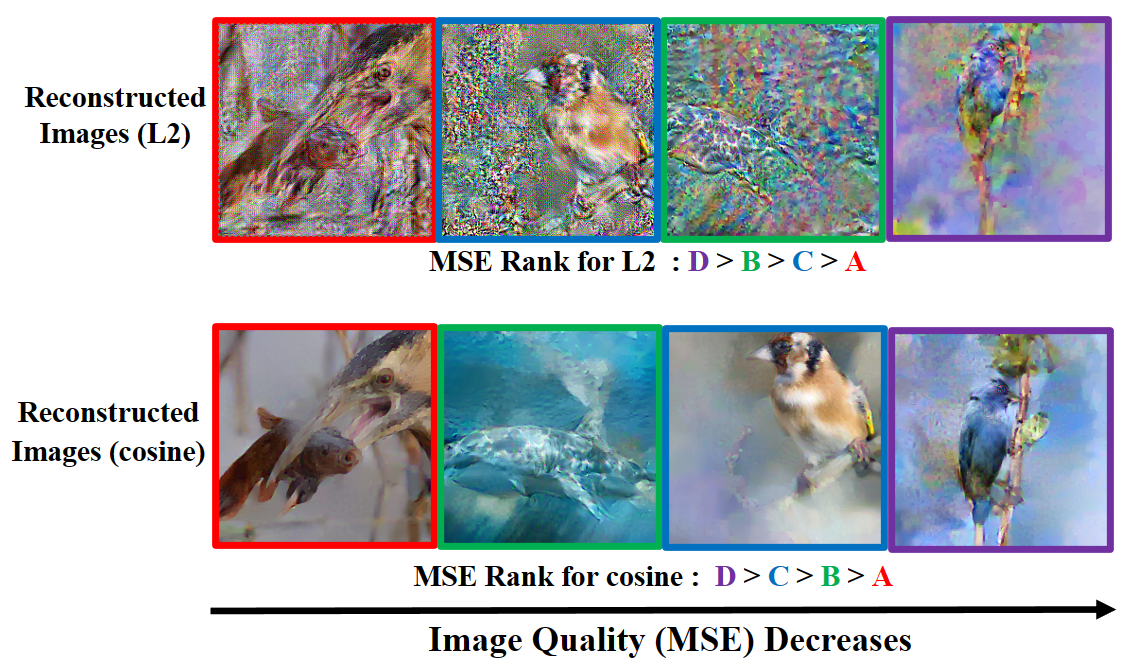}
        \caption{MSE ranks of reconstructed images from gradient inversion attacks with L2 and cosine distances}
        \label{fig:c3}
    \end{subfigure}
    \begin{subfigure}[b]{0.37\textwidth}
        \includegraphics[width=\textwidth]{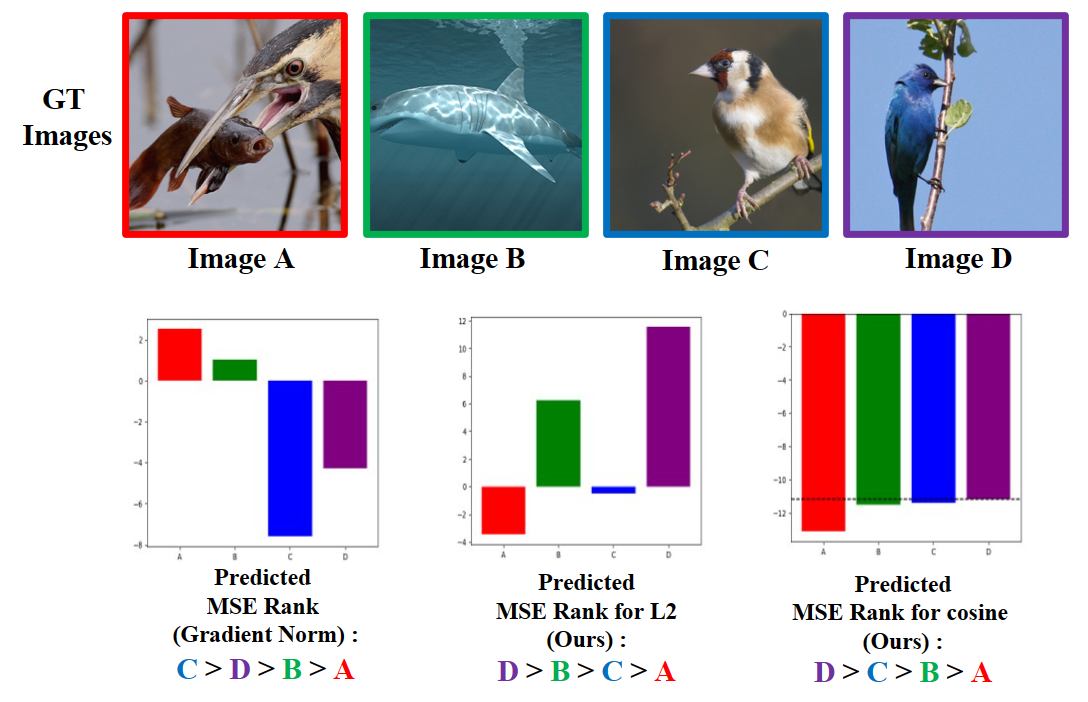}
        \caption{Predicted MSE ranks with gradient norm and LAVP (ours) for L2 and cosine distances }
        \label{fig:c2}
    \end{subfigure}
    \caption{ \textbf{Motivation of our work.} (a) A contemporary gradient inversion attack utilizes either L2 or cosine distance for gradient matching. (b) Distinct loss functions reveal different vulnerability rankings among images in Mean Squared Error (MSE). (c) We introduce a loss-aware vulnerability proxy (LAVP), capable of elucidating such loss-specific behaviors. LAVP for L2 and cosine distances predict MSE ranking \(\text{D} >\text{B} > \text{C} > \text{A}\) and \(\text{D} > \text{C} > \text{B} > \text{A}\), respectively. Each predicted ranking coincides with the correct MSE ranking in (b). In contrast, gradient norm, which remains constant regardless of the chosen loss functions cannot explain this.}
    \label{fig:concept}
\end{figure*}

\begin{abstract}

Gradient inversion attacks can leak data privacy when clients share weight updates with the server in federated learning (FL). Existing studies mainly use L2 or cosine distance as the loss function for gradient matching in the attack. Our empirical investigation shows that the vulnerability ranking varies with the loss function used. Gradient norm, which is commonly used as a vulnerability proxy for gradient inversion attack, cannot explain this as it remains constant regardless of the loss function for gradient matching. In this paper, we propose a loss-aware vulnerability proxy (LAVP) for the first time. LAVP refers to either the maximum or minimum eigenvalue of the Hessian with respect to gradient matching loss at ground truth. This suggestion is based on our theoretical findings regarding the local optimization of the gradient inversion in proximity to the ground truth, which corresponds to the worst case attack scenario. We demonstrate the effectiveness of LAVP on various architectures and datasets, showing its consistent superiority over the gradient norm in capturing sample vulnerabilities. The performance of each proxy is measured in terms of Spearman's rank correlation with respect to several similarity scores. This work will contribute to enhancing FL security against any potential loss functions beyond L2 or cosine distance in the future.

\end{abstract}

\section{Introduction}
\label{sec:intro}

Federated learning (FL) is a collaborative machine learning paradigm in which local clients act as trainers and a central server acts as a global aggregator~\cite{konevcny2016federated, mcmahan2017communication}. 
Each learning round in FL begins with the server distributing global model weights to participating clients.
Then, the clients compute weight updates for the shared global model based on their own data and send these updates back to the server. 
At the end of the round, the server aggregates all the weight updates received from participating clients for the update of global model.

An important aspect of FL is that participants cannot access the raw data of others, thus their communication is limited to exchanging weight updates.
These weight updates were previously believed to reveal minimal information about the original data.
However, recent studies~\cite{zhu2019deep, zhu2020r, geiping2020inverting, yin2021see, jeon2021gradient, kariyappa2023cocktail, zhu2023surrogate} have challenged this belief regarding data privacy in FL. 
They have demonstrated the possibility of an honest-but-curious server launching a gradient inversion attack, thereby stealthily recovering clients' data using weight gradients shared from clients.

In these attack algorithms, a randomly initialized input variable is optimized to match the current weight gradient computed with itself with the gradient shared from a client.
As a loss function for gradient matching, the literature primarily employs either \textit{L2 distance}~\cite{zhu2019deep, yin2021see} or \textit{cosine distance}~\cite{geiping2020inverting, jeon2021gradient, zhu2023surrogate} as in Figure~\ref{fig:c1}.

However, the reconstruction behavior of gradient inversion attack depends on the loss function for gradient matching. 
In Figure~\ref{fig:c3}, the L2 distance achieves a more accurate reconstruction for Image C (blue) than for Image B (green), while the cosine distance displays the opposite pattern.
The choice of loss function for gradient matching has a significant impact on the vulnerability ranking.

The gradient norm, commonly used as a vulnerability proxy in existing literature~\cite{geiping2020inverting, yin2021see}, remains constant regardless of the loss function for gradient matching.
Thus, it cannot account for the loss function dependence of vulnerability rankings among samples as described in~\ref{fig:c2}.
To address this issue, there is a need for a proxy that can provide a comprehensive explanation for the dependence of vulnerability rankings on the loss function.


In this paper, we introduce a novel loss-aware vulnerability proxy (LAVP) for the first time.
In specific, LAVP refers to either the maximum or minimum eigenvalue of the Hessian of gradient matching loss at the ground truth.
LAVP is founded on two theorems we have developed concerning gradient matching optimization. 
We prove that the gradient matching loss drops more significantly when bi-Lipschitz constants of the gradient function are smaller.
For simplicity, we focus on the local optimization near the ground truth, representing the worst-case attack scenario.
In this case, bi-Lipschitz constants near ground truth correspond to the maximum and minimum eigenvalues of the Hessian at the ground truth, which is how LAVP is derived.



We empirically show the efficacy of LAVP by presenting stronger correlation than the gradient norm, with the quality of reconstructed images from gradient inversion attacks. For both L2 and cosine distances, the vulnerability ranking among samples predicted by LAVP coincides better with the correct one than that predicted by the gradient norm as in Figure~\ref{fig:c2}. The superiority of LAVP over the gradient norm is consistently verified by experiments on diverse architectures and datasets ranging from low-resolution images in CIFAR-10, CIFAR-100, ImageNette, and ImageWoof, to high-resolution images in ImageNet.

The contribution of our work can be summarized as follows:

\begin{itemize}
    \item We propose using either the maximum or minimum eigenvalue of the Hessian at the ground truth as a loss-aware vulnerability proxy (LAVP) for the first time.

    \item We establish several theoretical results regarding the optimization of gradient inversion attacks in close proximity to the ground truth for the derivation of LAVP.

    \item We demonstrate the efficacy of LAVP in capturing vulnerability against gradient inversion attacks by comparing it to the gradient norm by thorough experiments.

    \item We propose the geometric mean between LAVP for L2 and cosine distances as the loss-agnostic proxy that caters to both L2 and cosine distances at once.
\end{itemize}


\section{Preliminaries: Gradient Inversion Attack}
\label{sec:preliminaries}

\subsection{Attack Scenario}

In a FL scenario, we assume that the server sends the global model \(f_w : \mathbb{R}^{b \times d} \rightarrow \mathbb{R}^{b \times c}\) to participating clients, where \(w\), \(b\), \(d\), and \(c\) denotes model weights, batch size, image size, and the number of classes, respectively. Subsequently, a client computes the weight gradient \(g^{*} = \frac{\partial \mathcal{L}(f_w(x^{*}), y^{*})}{\partial w}\) with respect to the private data batch \((x^{*}, y^{*}) \in \mathbb{R}^{b \times d} \times \mathbb{R}^{b}\) (\(x^{*}\) and \(y^{*}\) being image and label batches) using the objective function \(\mathcal{L} : \mathbb{R}^{b \times c} \times \mathbb{R}^{b} \rightarrow \mathbb{R}\). Then, the computed gradient is sent back to the server. In this setup, the server, acting as an honest-but-curious adversary, could attempt to reconstruct an image batch \(x \in \mathbb{R}^{b \times d}\) resembling the ground truth batch \(x^{*}\), using the available information \(g^{*}\) and \(f_w\). \textit{For brevity, we assume \(b = 1\) throughout the paper.}

\subsection{Optimization based Gradient Inversion Attacks}

Gradient inversion attacks aim to reconstruct input batch by minimizing the distance between the current gradients and the target gradients as follows:
\begin{equation}
    \label{GradInv1}
    \argminA_{x, y} \mathcal{L}_{grad}( g_w(x, y) , g^{*} )
    + \alpha_{prior} \mathcal{R}_{prior} (x),
\end{equation}
where \(g_w(x, y) = \frac{\partial \mathcal{L}(f_w(x), y)}{\partial w}\) represents the weight gradient as a function of the input batch and \(\mathcal{L}_{grad} : \mathbb{R}^{n} \times \mathbb{R}^{n} \rightarrow \mathbb{R}\) (\(n\) is the dimension of model weights \(w\)) serves as the loss function for gradient matching. Also, \(\mathcal{R}_{prior} : \mathbb{R}^{b \times d} \rightarrow \mathbb{R}\) denotes the regularization loss for image prior and \(\alpha_{prior}\) represents its coefficient.
Especially, gradient matching loss function $\mathcal{L}_{grad}$ is chosen to cosine distance ($\mathcal{L}_{grad}(g, g^{*}) = 1 - \frac{<g, g^{*}>}{||g|| ||g^{*}||}$)~\cite{geiping2020inverting, jeon2021gradient, huang2021evaluating, yin2021see, zhu2023surrogate} or L2 distance ($\mathcal{L}_{grad}(g, g^{*}) = ||g - g^{*}||_{2}^{2}$)~\cite{zhu2019deep, zhao2020idlg, yin2021see}.

  

\subsection{Enhanced Assumptions for Stronger Attacks}  

Beyond the baseline attack, which solely relies on gradient matching, recent gradient inversion attack methods introduce several augmented assumptions for stronger attack.

Firstly, it is assumed that the server knows private labels associated with clients' data. Recent works solve the optimization problem presented in Equation (\ref{GradInv1}) in a sequential manner. This involves initially estimating the labels \(y\) directly through \(g^{*}\) and \(f_w\)~\cite{ma2022instance, wainakh2021label, zhao2020idlg, yin2021see}, followed by exclusive optimization of \(x\) using Equation (\ref{GradInv1}), drawing upon the earlier approximated \(y = y^{*}_{approx}\). This disentangles label estimation from the optimization problem in Equation (\ref{GradInv1})~\cite{dang2021revealing, ye2022label, li2021label}. Consequently, recent studies have predominantly focused on image reconstruction under the premise of private label knowledge~\cite{geiping2020inverting, jeon2021gradient}. \textit{We also embrace this assumption in our work.}

Secondly, local batch statistics $\{ {\mu_l^{*}}, {\sigma_l^{*}}^{2}  \} _{l=1}^{N} $ are computed with clients' data batch and then shared with the server alongside weight updates, where ${ \mu_l^{*}}$, ${{\sigma_l^{*}}^{2}}$, and $N$ signify batch mean, batch variance, and the count of batch normalization (BN) layers, respectively. Utilizing local batch statistics indeed contributes to the reconstruction of high-resolution images (with batch size up to 40) of superior quality~\cite{yin2021see, hatamizadeh2022gradient}. However, the sharing of batch statistics is not a mandatory requirement for clients, thus \textit{we reject this assumption}.

  
\section{Related Work: Proxies for the Vulnerability against Gradient Inversion Attack}
\label{sec:related_work}
\noindent \textbf{Gradient Norm.} 
In recent studies~\cite{yin2021see, geiping2020inverting}, the gradient norm is frequently employed as a heuristic proxy for vulnerability assessment against gradient inversion attacks. This approach is grounded in the intuition that a gradient norm close to zero implies negligible information, hence leading to reconstruction failure. In \cite{yin2021see}, the proposed metric for batch reconstruction, termed Image Identifiability Precision (IIP) is demonstrated on images with higher gradient norms that are perceived as more susceptible examples to gradient inversion attacks. Furthermore, \cite{geiping2020inverting} introduces a label flipping attack, which pertains to permuting classifier weights rather than label inversion. To address concerns that a fully trained classifier might yield lower-norm gradients, a threat model is established wherein a malicious server swaps the classifier channel for the correct label with that for any incorrect label. \textit{However, the gradient norm lacks theoretical or empirical foundation as a vulnerability proxy.} 

\noindent \textbf{Jacobian Norm.} The utilization of the Jacobian norm as a proxy to quantify the extent of input information within gradients was explored in previous work~\cite{moquantifying}. Employing usable information theory, the sensitivity, denoted as $E_{\Delta x} [||g_{w}(x^{*} + \Delta x) - g^{*}||]$, is interpreted as an indicator of input information contained within gradients. The sensitivity is reformulated into the Jacobian norm in~\cite{moquantifying}, making it the most closely aligned with LAVP (ours) for the L2 distance metric (the maximum eigenvalue of the Jacobian) among preceding proxies. Note that the maximum eigenvalue of the Jacobian corresponds to the spectral norm of the Jacobian. \textit{However, the interpretation of the Jacobian norm fundamentally differs from our perspective.} In~\cite{moquantifying}, higher sensitivity of the gradient around the ground truth suggests that the gradient is more likely to be unique within the vicinity of the ground truth, thus making it more susceptible to revealing input information. In contrast, from our optimization viewpoint, a greater gradient sensitivity indicates convergence instability, making optimization of the gradient matching loss more challenging. Indeed, our experimental results align with this intuition. In addition, the objective of~\cite{moquantifying} is to identify layers in which the gradient component significantly encodes input information. We focus on a sample-wise approach rather than a layer-wise approach (i.e., identifying vulnerable examples, not layers). This explains why~\cite{moquantifying} is not regarded as a competitor to our proposed method.
\section{Method}
\label{sec:method}
In this section, we present a novel loss-aware vulnerability proxy called LAVP to effectively elucidate loss-specific reconstruction behaviors of gradient inversion attacks. We claim that the susceptibility to the gradient inversion attack is inversely proportional to the bi-Lipschitz constants of the gradient function $g_w$, denoted as $L$ and $M$. This claim is backed by our proofs of two theorems regarding $L$ and $M$ respectively. We establish a correspondence of $L$ and $M$ near ground truth to the maximum and minimum eigenvalues of the Hessian with respect to the gradient matching loss \(\mathcal{L}_{grad}\) at ground truth $x^{*}$. In the end, we outline the methodology for computing both maximum and minimum eigenvalues of Hessian, for both L2 and cosine distances.

\subsection{Theoretical Results on the Optimization of Gradient Matching}
If a function $f : \mathbb{R}^n \rightarrow \mathbb{R}^m$ is Lipschitz continuous with constant $L$, then the following holds: $||f(x) - f(y)|| \leq L ||x - y|| \; \forall x, y \in \mathbb{R}^n$. We employ the concept of Lipschitz continuity to prove the following theorem in the context of gradient matching problem. Note that we use $g_w (x)$ instead of $g_w (x, y)$ throughout this section by the aforementioned assumption that label information is available.

\begin{theorem}
\normalfont (The first theorem on gradient matching optimization). Suppose $g_w(x)$ is Lipschitz continuous with respect to $x$ with constant $L$ and $\mathcal{L}_{grad}(x) = ||g_w(x) - g^{*}||_2^2 $ is the gradient matching loss. Then, when gradient descent $\Delta x$ is applied with step size $\mu = \frac{1}{2L^2} > 0$ and $L > \epsilon$ for some $\epsilon > 0$, the following holds:

\begin{equation}
    \label{dl-ineq}
    \mathcal{L}_{grad}(x+ \mu\Delta x) \leq \mathcal{L}_{grad}(x) - \frac{1}{4L^2} || \frac{\partial \mathcal{L}_{grad}(x)}{\partial x}||_{2}^{2},
\end{equation}
where $L > \epsilon$ satisfies $|| \mu \Delta x|| < \delta$ such that $g_w(x + \mu \Delta x) = g_w(x) + \mu \nabla_x g_w(x)^{T} \Delta x$ holds approximately.
\label{monotonic}
\end{theorem}

\begin{proof}
See Appendix A1.
\end{proof} 

Inequality (\ref{dl-ineq}) implies that gradient matching loss strictly decreases as the gradient descent steps unless the gradient term $\frac{\partial \mathcal{L}_{grad}(x)}{\partial x}$ is zero (i.e. gradient matching loss already converges). Then, a gradient descent with a small $L$ can accelerate the convergence of gradient matching optimization. Instead, there is the premise that $L>\epsilon$ for $\epsilon > 0$, which is required for Taylor's first approximation on $g_w(x)$. Therefore, in a particular range of $L$ (i.e., $L > \epsilon$), we hypothesize that a global model with smaller $L$ experiences a sharper loss drop in gradient matching optimization. 

In Theorem~\ref{monotonic}, optimal loss drop is achieved when $\mu = \frac{1}{2L^2}$. In the proof of Theorem~\ref{monotonic}, $\mu$ should be the minimizer of the last term on the right hand for optimal loss drop, while there would be no such restriction if the term was on the left hand. Therefore, if there is a Lipschitz constant for opposite direction denoted by $M > 0$ such that  $|| g_w(x_1) - g_w(x_2)|| \geq M || x_1 - x_2|| \forall x_1, x_2$), the following theorem can be derived.

\begin{theorem}
\normalfont (The second theorem on gradient matching optimization). Suppose $|| g_w (x_1) - g_w (x_2)|| \geq M || x_1 - x_2|| \forall x_1, x_2$ for $M > 0$ holds and $\mathcal{L}_{grad}(x) = ||g_w(x) - g^{*}||_2^2 $ is the gradient matching loss. Then, when gradient descent $\Delta x$ is applied with step size $\mu  < \delta_1$ for some $\delta_1 > 0$, the following holds:

\begin{equation}
    \label{dl-ineq2}
    \mathcal{L}_{grad}(x + \mu\Delta x) \geq \mathcal{L}_{grad}(x) - \frac{1}{4M^2} ||\frac{\partial \mathcal{L}_{grad}(x)}{\partial x}||_{2}^{2},
\end{equation}
where $\mu < \delta_1$ satisfies $|| \mu \Delta x|| < \delta$ such that $g_w (x + \mu \Delta x) = g_w(x) + \mu \nabla_x g_w(x)^{T} \Delta x$ holds approximately.
\label{monotonic2}
\end{theorem}

\begin{proof}
See Appendix A2.
\end{proof}

The proof of Theorem~\ref{monotonic2} is similar to that of Theorem~\ref{monotonic} except that the term including $\mu$ to be minimized is on the left side, thus there is no restriction like $\mu = \frac{1}{2L^2}$ in Theorem~\ref{monotonic}, thus more favorable. Inequality (\ref{dl-ineq2}) implies that the upper bound of gradient matching loss drop is $\frac{1}{4M^2} ||\frac{\partial \mathcal{L}_{grad}(x)}{\partial x}||_{2}^{2}$, unless the gradient term $\frac{\partial \mathcal{L}_{grad}(x)}{\partial x}$ is zero. Then, a gradient descent with a large $M$ can hinder the convergence of gradient matching optimization. Therefore, we hypothesize that a global model with smaller $M$ has a potential to experience a sharper loss drop in gradient matching optimization. 

\subsection{Finding $L$ and $M$ near Ground Truth: Maximum and Minimum Eigenvalues of Hessian}
Theorems~\ref{monotonic} and~\ref{monotonic2} are about one-step loss drop, so summarizing the whole process of optimization with them is difficult. To mitigate such complexity, we consider the loss drop near ground truth as it is the most important to decide whether ground truth can be reached through optimization or not. For a neighborhood point of of ground truth $x^{*}$, $x^{*} + \Delta x$ ($||\Delta x||$ is very small), gradient matching loss can be approximated by Taylor's second-order approximation like the following: 

\begin{align*}
      \mathcal{L}_{grad}(x^{*} + \Delta x) = &\mathcal{L}_{grad}(x^{*}) + \nabla_{x = x^{*}} \mathcal{L}_{grad}(x)^{T}\Delta x +\\ &\frac{1}{2} \Delta x^{T}H(x^{*})\Delta x, 
\end{align*}
where $H(x^{*})$ is the Hessian of gradient matching loss with respect to input variables at $x^{*}$. Note that both $ \mathcal{L}_{grad}(x^{*})$ and $\nabla_{x=x^{*}} \mathcal{L}_{grad}(x)$ are zero when $\mathcal{L}_{grad}$ is either L2 or cosine distance. Then, $\mathcal{L}_{grad}(x^{*} + \Delta x)$ can be interpreted as the distance in gradient space while $||\Delta x||^2$ corresponds to distance in input space. Combining two preceding observations, the ratio of gradient distance to input distance is $\mathcal{L}_{grad}(x^{*}+\Delta x)/||\Delta x||^2 = \frac{1}{2}\frac{\Delta x}{||\Delta x||}^{T}H(x^{*})\frac{\Delta x}{||\Delta x||}$. Then, the upper and lower bounds of this ratio correspond to maximum and minimum eigenvalues of Hessian, respectively. In proximity to ground turth, we can replace $L$ and $M$ with maximum and minimum eigenvalues of Hessian at ground truth, which is our proposed proxy, LAVP.

\begin{table*}[t]
\centering
\resizebox{2\columnwidth}{!}{
\begin{tabular}{|c|c|c|c|c|c|}
\hline
$\sigma_{S}$           & grad\_norm  & max (LAVP for L2)         & min (LAVP for L2)         & ang\_max (LAVP for CS)  & ang\_min (LAVP for CS)    \\ \hline
C-10+L2   & 0.35 / -0.27 / -0.35 / -0.13   &  \textbf{0.51} / \textbf{-0.46} / \textbf{-0.51} / -0.15  & 0.41 / -0.40 / -0.41 / \textbf{-0.20} & -0.06 / -0.01 / 0.06 / -0.06    & -0.04 / -0.07 / 0.04 / -0.06    \\ \hline
C-100+L2  & 0.41 / -0.31 / -0.41 / -0.19  & \textbf{0.46} / \textbf{-0.57} / \textbf{-0.46} / 0.10    & 0.41 / -0.45 / -0.41 / -0.01   & -0.05 / -0.18 / 0.05 / \textbf{0.22}    & -0.05 / -0.20 / 0.05 / 0.19    \\ \hline
IN+L2 & 0.03 / 0.03 / -0.03 / 0.19 & 0.33 / -0.26 / -0.33 / \textbf{0.49}  & \textbf{0.34} / -0.25 / \textbf{-0.35} / 0.46 & 0.14 / -0.20 / -0.14 / 0.42   & 0.25 / \textbf{-0.34} / -0.25 / 0.42  \\ \hline
IW+L2 & 0.35 / -0.01 / -0.35 / 0.00 & 0.66 / \textbf{-0.58} / -0.66 / 0.46  & \textbf{0.68} / -0.52 / \textbf{-0.68} / 0.29 & 0.38 / -0.41 / -0.38 / 0.46   & 0.46 / -0.52 / -0.46 / \textbf{0.49}  \\ \hline
C-10+CS   & -0.28 / 0.25 / 0.28 / -0.18   & -0.03 / -0.02 / 0.03 / 0.00   & 0.04 / -0.08 / -0.04 / 0.02  & 0.26 / -0.31 / -0.26 / 0.36  & \textbf{0.64} / \textbf{-0.74} / \textbf{-0.64} / \textbf{0.62}  \\ \hline
C-100+CS  & 0.10 / -0.07 / -0.10 / 0.02   & 0.37 / -0.35 / -0.37 / 0.26   & 0.32 / -0.34 / -0.32 / 0.24 & 0.67 / -0.8 / -0.67 / \textbf{0.68}  & \textbf{0.69} / \textbf{-0.81} / \textbf{-0.69} / 0.65   \\ \hline
IN+CS  & -0.13 / 0.11 / 0.14 / -0.18   & 0.16 / -0.16 / -0.16 / 0.12   & 0.27 / -0.28 / -0.25 / 0.17  & 0.57 / -0.65 / -0.57 / 0.64  & \textbf{0.75} / \textbf{-0.80} / \textbf{-0.75} / \textbf{0.73}   \\ \hline
IW+CS  & 0.00 / 0.01 / 0.00 / -0.04   & 0.37 / -0.37 / -0.37 / 0.21   & 0.33 / -0.34 / -0.33 / 0.18  & 0.72 / -0.73 / -0.72 / \textbf{0.61}  & \textbf{0.75} / \textbf{-0.83} / \textbf{-0.75} / 0.61   \\ \hline

\end{tabular}
}
\vspace{-7pt}
\caption{\textbf{Spearman's correlation ($\sigma_{S}$) of proxy candidates with image similarity scores (MSE($\downarrow$) / SSIM($\uparrow$) / PSNR($\uparrow$) / LPIPS($\downarrow$)) on low-resolution images.} 'C-10', 'C-100', 'IN', and 'IW' denote CIFAR-10, CIFAR-100, ImageNette, and ImageWoof respectively. 'L2' and 'CS' denote L2 and cosine distances respectively. 'grad\_norm', 'max', 'min', 'ang\_max', and 'ang\_min' denote the gradient norm, the maximum eigenvalue of Hessian for L2, the minimum eigenvalue of Hessian for L2, the maximum eigenvalue of Hessian for CS, and the minimum eigenvalue of Hessian for CS. }
\label{tab:corr-low}
\end{table*}

\begin{figure*}[thb!]

\begin{subfigure}{.325\textwidth}
  \centering
  \includegraphics[width=1\linewidth]{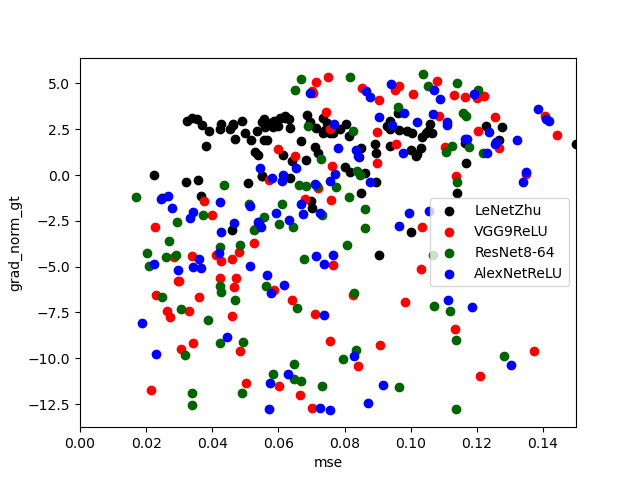}  
  \caption{ grad\_norm vs MSE (L2, $\sigma_{S} = 0.35$)}
  \label{fig:mse-gn-l2-cifar10}
\end{subfigure}
\begin{subfigure}{.325\textwidth}
  \centering
  \includegraphics[width=1\linewidth]{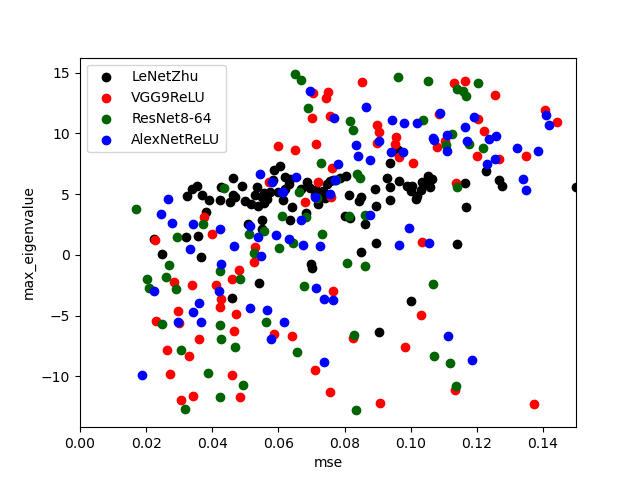}
  \caption{max vs MSE (L2, $\mathbf{\sigma_{S} = 0.51}$)}
  \label{fig:mse-max-l2-cifar10}
\end{subfigure}
\begin{subfigure}{.325\textwidth}
  \centering
  \includegraphics[width=1\linewidth]{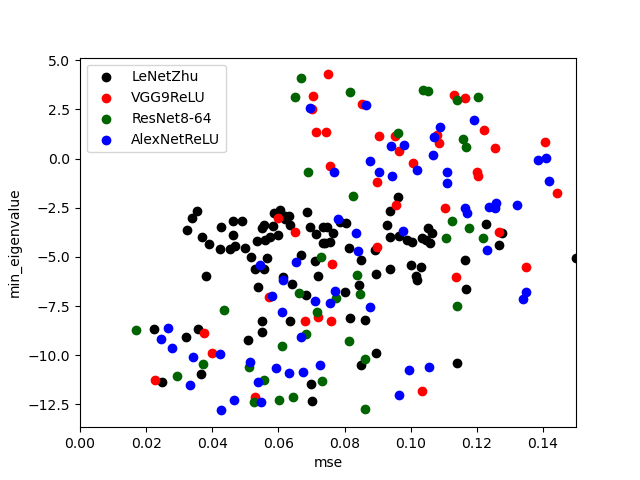}  
  \caption{min vs MSE (L2, $\sigma_{S} = 0.41$)}
  \label{fig:mse-min-l2-cifar10}
\end{subfigure}
\begin{subfigure}{.325\textwidth}
  \centering
  \includegraphics[width=1\linewidth]{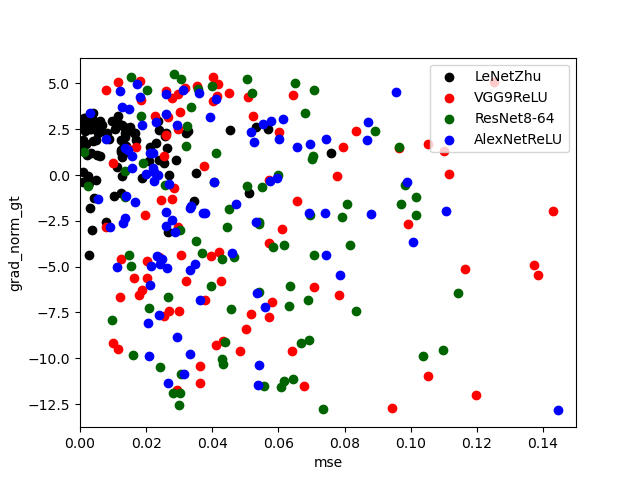}  
  \caption{grad\_norm vs MSE (CS, $\sigma_{S} = -0.28$)}
  \label{fig:mse-gn-sim-cifar10}
\end{subfigure}
\begin{subfigure}{.325\textwidth}
  \centering
  \includegraphics[width=1\linewidth]{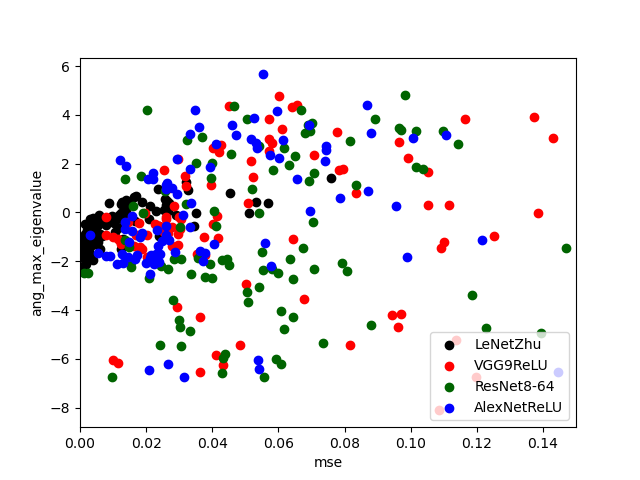}  
  \caption{ang\_max vs MSE (CS, $\sigma_{S} = 0.26$)}
  \label{fig:mse-angmax-sim-cifar10}
\end{subfigure}
\begin{subfigure}{.325\textwidth}
  \centering
  \includegraphics[width=1\linewidth]{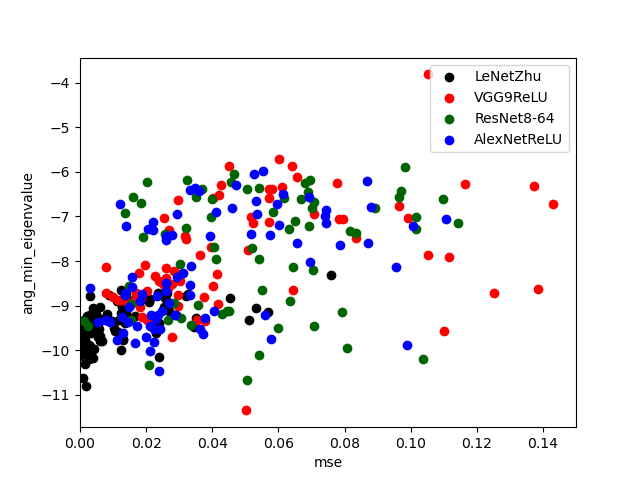}  
  \caption{ang\_min vs MSE (CS,  $\mathbf{\sigma_{S} = 0.64}$))}
  \label{fig:mse-angmin-sim-cifar10}
\end{subfigure}

\caption{\textbf{Comparison of the gradient norm, maximum and minimum eigenvalues of Hessian in terms of the correlation with MSE of reconstructed samples over several architectures on CIFAR10 test samples.}}
  
\label{fig:corr-low}
\end{figure*}

\subsection{Hessian of Gradient Matching Loss}
To compute LAVP, Hessian should be identified first. In Theorems~\ref{hes-l2} and~\ref{hes-cos}, we derive the Hessian for L2 and cosine distances in a closed form respectively. 

\begin{theorem}
\normalfont (Hessian at ground truth for L2 distance). Suppose $\mathcal{L}_{grad}$ is L2 distance, then the Hessian at ground truth is like the following:

\begin{equation}
    \label{hes-l2}
    H_{\text{L2}}(x^{*}) = J(x^{*})^{T}J(x^{*}),
\end{equation}
where $J(x^{*}) = \nabla_{x=x^{*}}g_w(x)$ is the Jacobian of gradient function $g_w(x)$ with respect to input at ground truth $x^{*}$.
\label{hessian-l2}
\end{theorem}

\begin{proof}
See Appendix A3.
\end{proof}

When $\mathcal{L}_{grad}$ is L2 distance, $H_{\text{L2}}(x^{*}) = J(x^{*})^{T}J(^{*})$ holds by Theorem~\ref{hessian-l2}, thus positive semi-definite. since input dimension is smaller than weight dimension in general, $H_{\text{L2}}(x^{*})$ is not trivial low rank and its minimum eigenvalue has a potential to be positive. 

For cosine distance, Hessian at ground truth can be solved in closed form by the following theorem.

\begin{theorem}
\normalfont (Hessian at ground truth for cosine distance). Suppose $\mathcal{L}_{grad}$ is cosine distance, then the Hessian at ground truth is like the following:

\begin{equation}
    \label{hes-cos}
    H_{\text{cos}}(x^{*}) = \frac{1}{||g^{*}||^2}J(x^{*})^{T}(I - \frac{g^{*}}{||g^{*}||}\frac{g^{*T}}{||g^{*}||})J(x^{*}),
\end{equation}
where $I$ is the identity matrix.
\label{hessian-cos}
\end{theorem}

\begin{proof}
See Appendix A4.
\end{proof}

The minimum eigenvalue of $H_{\text{cos}}(x^{*})$ is nonnegative as it is positive semi-definite by Cauchy-Schwartz inequality. 




\subsection{Implementation of LAVP}
To find the maximum eigenvalue of Hessian, power iteration is used. Power iteration computes matrix-vector product and normalization alternatively until the vector converges to the eigenvector with the maximum eigenvalue. When this algorithm is applied to the Hessian, Jacobian-vector product is inevitable, while \textit{Autograd} package in PyTorch supports only vector-Jacobian product. Therefore, Jacobian-vector product is solved with the finite difference method with very small step size. Once the maximum eigenvalue $\mathcal{\alpha}_{max}$ is obtained for the Hessian $H(x^{*})$, then power iteration is applied to $\alpha_{max}I - H(x^{*})$ ($I$ is the identity matrix) for identifying the minimum eigenvalue $\alpha_{min}$, as $\alpha_{max}-\alpha_{min}$ would be the maximum eigenvalue of $\alpha_{max}I - H(x^{*})$ ($I$ is the identity matrix). \textit{It is noteworthy that multiple Hessian-vector products, rather than the entire Hessian, are sufficient for computing LAVP, thus more efficient.}

\begin{table*}[t]
\centering
\resizebox{2\columnwidth}{!}{
\begin{tabular}{|c|c|c|c|c|c|}
\hline
$\sigma_{S}$           & grad\_norm  & max (LAVP for L2)         & min (LAVP for L2)        & ang\_max (LAVP for CS)  & ang\_min (LAVP for CS)   \\ \hline

ImageNet+L2 & 0.66 / -0.66 / -0.66 / \textbf{-0.28} & 0.69 / -0.72 / -0.69 / -0.09  & \textbf{0.74} / \textbf{-0.74} / \textbf{-0.72} / -0.07 & -0.05 / 0.09 / 0.05 / 0.03   & -0.07 / 0.12 / 0.07 / 0.10  \\ \hline

ImageNet+CS  & -0.06 / -0.04 / 0.00 / -0.05  & 0.02 / -0.07 / -0.11 / 0.04   & -0.03 / -0.05 / -0.06 / 0.00  & \textbf{0.27} / \textbf{-0.37} / -0.21 / 0.32  & 0.26 / -0.22 / \textbf{-0.24} / \textbf{0.32}   \\ \hline

\end{tabular}
}
\vspace{-7pt}
\caption{\textbf{Spearman's correlation of proxy candidates with image similarity scores (MSE($\downarrow$) / SSIM($\uparrow$) / PSNR($\uparrow$) / LPIPS($\downarrow$)) on ImageNet.\vspace*{-5mm}} }
\label{tab:corr-high}
\end{table*}

\begin{figure*}[h]

  \begin{subfigure}{.325\textwidth}
  \centering
  \includegraphics[width=1\linewidth]{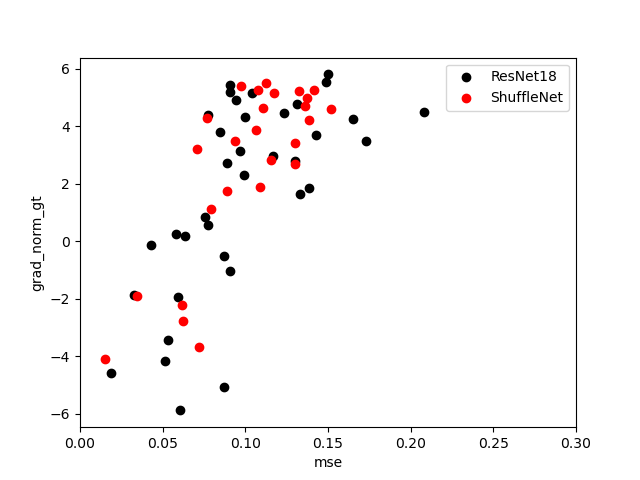}  
  \caption{ grad\_norm vs MSE (L2, $\sigma_{S} = 0.66$)}
  \label{fig:mse-gn-l2-imagenet}
\end{subfigure}
\begin{subfigure}{.325\textwidth}
  \centering
  \includegraphics[width=1\linewidth]{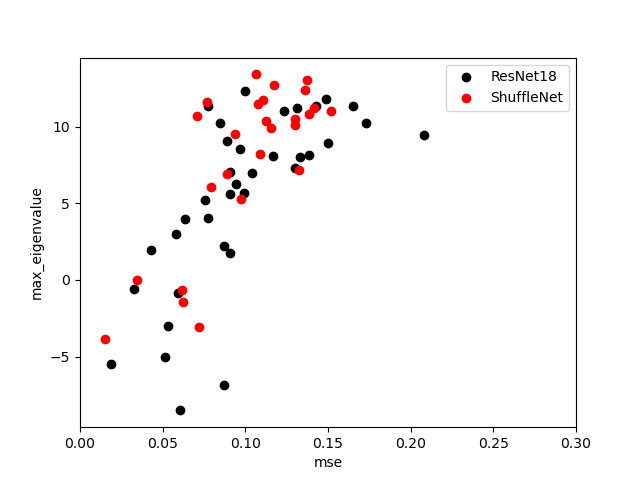}
  \caption{max vs MSE (L2, $\sigma_{S} = 0.69$)}
  \label{fig:mse-max-l2-imagenet}
\end{subfigure}
\begin{subfigure}{.325\textwidth}
  \centering
  \includegraphics[width=1\linewidth]{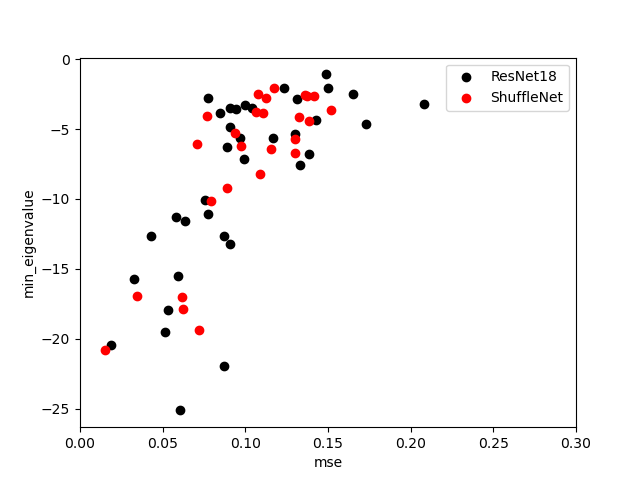}  
  \caption{min vs MSE (L2, $\mathbf{\sigma_{S} = 0.74}$)}
  \label{fig:mse-min-l2-imagenet}
\end{subfigure}
\begin{subfigure}{.325\textwidth}
  \centering
  \includegraphics[width=1\linewidth]{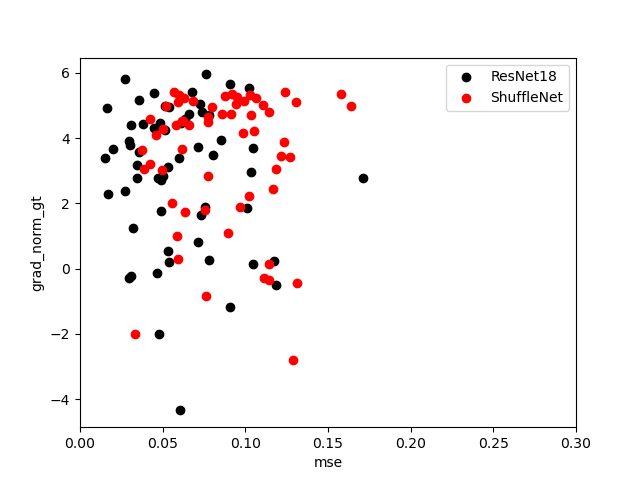}  
  \caption{grad\_norm vs MSE (CS, $\sigma_{S} = -0.06$)}
  \label{fig:mse-gn-sim-imagenet}
\end{subfigure}
\begin{subfigure}{.325\textwidth}
  \centering
  \includegraphics[width=1\linewidth]{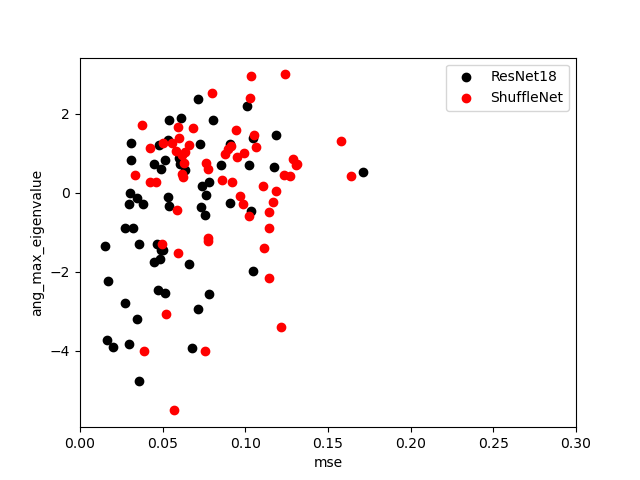}  
  \caption{ang\_max vs MSE (CS, $\mathbf{\sigma_{S} = 0.27}$)}
  \label{fig:mse-angmax-sim-imagenet}
\end{subfigure}
\begin{subfigure}{.325\textwidth}
  \centering
  \includegraphics[width=1\linewidth]{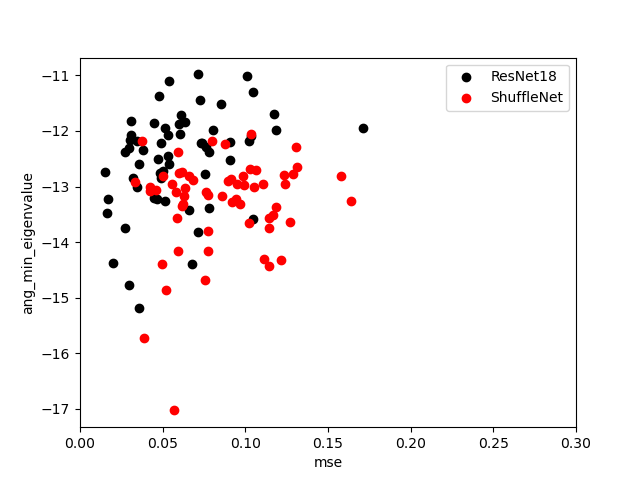}  
  \caption{ang\_min vs MSE (CS, $\sigma_{S} = 0.26$))}
  \label{fig:mse-angmin-sim-imagenet}
\end{subfigure}

\caption{\textbf{Comparison of the gradient norm, maximum and minimum eigenvalues of 
Hessian in terms of the correlation with MSE of reconstructed samples over ResNet18 and ShuffleNet models on ImageNet validation samples.\vspace*{-3mm}}}
  
\label{fig:corr-high}
\end{figure*}

\section{Experimental Results}
\label{sec:experiment}
In this section, we begin with a concise overview of our experimental setup. Then, we elucidate the advantages of LAVP over the gradient norm (baseline), in explaining the vulnerability to the gradient inversion attack with either L2 or cosine distance by providing correlation tables and plots.
For a black-box scenario where the attacker's loss function is unknown to clients, we also introduce the loss-agnostic LAVP fusion, the proxy that can handle several candidate loss functions at once. An example of LAVP fusion includes the geometric mean between LAVPs for L2 and cosine distances. We provide the correlation table for this example with a comparative evaluation against the gradient norm.

\begin{table}[h]
\centering
\resizebox{0.9\columnwidth}{!}{
\centering
\begin{tabular}{|c|c|c|}

\hline
$\sigma_{S}$  &     grad\_norm   &     $\sqrt{\text{max * ang\_min}}$  (LAVP fusion)  \\ \hline
C-10+L2   & 0.35 / -0.27 / -0.35 / \textbf{-0.13}   &  \textbf{0.48} / \textbf{-0.49} / \textbf{-0.48} / -0.09   \\ \hline
C-100+L2  &0.41 / -0.31 / -0.41 / \textbf{-0.19}    &  \textbf{0.50} / \textbf{-0.49} / \textbf{-0.50} / -0.09 \\ \hline
IN+L2 &  0.03 / 0.03 / -0.03 / 0.19 & \textbf{0.48} / \textbf{-0.41} / \textbf{-0.48} / \textbf{0.59} \\ \hline
IW+L2 &  0.35 / -0.01 / -0.35 / 0.00 & \textbf{0.71} / \textbf{-0.71} / \textbf{-0.71} / \textbf{0.51} \\ \hline
C-10+CS   & -0.21 / \textbf{0.25} / -0.21 / 0.15   & \textbf{-0.28} / \textbf{0.25} / \textbf{0.28} / \textbf{-0.18} \\ \hline
C-100+CS  & 0.10 / -0.07 / -0.10 / 0.02  &  \textbf{0.50} / \textbf{-0.45} / \textbf{-0.50} / \textbf{0.36}  \\ \hline
IN+CS  &  -0.13 / 0.11 / 0.14 / -0.18  & \textbf{0.44} / \textbf{-0.43} / \textbf{-0.44} / \textbf{0.34}  \\ \hline
IW+CS  & 0.00 / 0.01 / 0.00 / -0.04  & \textbf{0.57} / \textbf{-0.59} / \textbf{-0.57} / \textbf{0.38} \\ \hline

\end{tabular}\par
}
\vspace{-6pt}
\caption{\textbf{Spearman's correlation of loss-agnostic LAVP and gradient norm with image similarity scores (MSE($\downarrow$) / SSIM($\uparrow$) / PSNR($\uparrow$) / LPIPS($\downarrow$)).} An instance of LAVP fusion is the geometric mean between the maximum eigenvalue of Hessian for L2 distance and the minimum eigenvalue of Hessian for cosine distance. }
\label{tab:loss-agnostic}
\end{table}

\subsection{Experimental Setup} \label{results:es}

\noindent \textbf{Datasets.} 
We conducted an evaluation by randomly selecting 100 validation images from CIFAR-10~\cite{alex2009learning}, CIFAR-100~\cite{alex2009learning}, ImageNette~\cite{imagewang}, ImageWoof~\cite{imagewang}, and ImageNet~\cite{deng2009imagenet}. Notably, ImageNette and ImageWoof are subsets of ImageNet~\cite{deng2009imagenet}, each consisting of ten easily classified classes, but with different classes from one another.

\noindent \textbf{Architectures and Attack Hyperparameters.}
We evaluated several deep learning models on low-resolution images, including LeNet~\cite{lecun1998gradient}, AlexNet~\cite{krizhevsky2017imagenet}, VGG9~\cite{simonyan2014very}, and ResNet8~\cite{he2016deep}. We trained these models on a training set for 300 epochs, using the SGD optimizer with an initial learning rate of 0.1 and a learning rate decay of 0.1 at the $150^{th}$ and $225^{th}$ epochs. We also trained ResNet18~\cite{he2016deep} and ShuffleNet~\cite{ma2018shufflenet} models on high-resolution images from ImageNet. To assess the vulnerability to attacks, we directly performed gradient inversion attacks on 100 validation images randomly selected from each dataset. We used attack algorithms from previous works~\cite{geiping2020inverting, yin2021see, zhu2019deep} and considered two major gradient matching losses: L2 and cosine distances. Also, we incorporated the total variation loss for regularization. We use Adam optimizer~\cite{kingma2015adam} for gradient inversion. For each sample, we run attack algorithm three times using different random seeds. The final outcome was recorded the one that reconstructs the best among these runs.

\noindent \textbf{Image Similarity Scores.}
 Image similarity scores measure the quality of reconstructed images compared to the original images. We consider Mean Squared Error (MSE), Learned Perceptual Image Patch Similarity (LPIPS)~\cite{zhang2018unreasonable}, Structural Similarity Index (SSIM)~\cite{wang2004image} and Peak Signal-to-Noise Ratio (PSNR) for quantifying reconstruction quality. MSE computes the mean pixel-wise difference between original sample and its reconstruction in image space. LPIPS computes the distance from ground truth within the feature space of the ImageNet-pretrained VGG network. SSIM measures the similarity between two images by comparing their structural information, luminance, and contrast. PSNR measures the quality of reconstructed images using signal-to-noise (SNR) ratio.




\noindent \textbf{Proxy for the Vulnerability.}
We consider the gradient norm and LAVP as the candidates for the proxy. In tables, the gradient norm is denoted as `grad$\_$norm', maximum and minimum eigenvalues for L2 distance hessian are denoted as `max' and `min', and maximum and minimum eigenvalues for cosine distance loss are denoted as `ang\_max' and `ang\_min'. Here, `ang' is the abbreviation for `angular'.

\noindent \textbf{Evaluation Metric.}
For a random variable of reconstructed images $X$, we compute the correlation between two mapping of $X$, $A(X)$ and $B(X)$ where A is a similarity score, and B is a proxy for the vulnerability. Pearson's correlation coefficient ($\sigma_{P}$) is often used to compute linear correlation, while monotonocity is more important than linearity in our case. Thus, we use Spearman's correlation coefficient ($\sigma_{S}$) to compute monotonic relationship between $A(X)$ and $B(X)$. Note that Spearman's correlation coefficient is Pearson's correlation between $Rank(A(X))$ and $Rank(B(X))$, where $Rank(\cdot)$ is the operator for ranking numbers in increasing order. The correlation is said to be strong when the absolute value of $\sigma_{S}$ is close to one. Specifically, intra correlation within the same architecture is more important as vulnerable examples might depend on the model. \textit{Note that $\sigma_{S}$ is computed for each architecture and their average is reported as the final evaluation metric.}

\subsection{Results for the Correlation between the Proxy and Vulnerability}

\label{results:cr1}
Table~\ref{tab:corr-low} and Table~\ref{tab:corr-high} present correlation results of proxy candidates on several combinations of dataset and loss function type for low resolution images and high resolution images. \textit{Note that the sign of the correlation depends on the image similarity score due to the fact that both MSE and LPIPS decrease as image quality improves, whereas the reverse is true for SSIM and PSNR.} When gradient inversion is based on the L2 (cosine) distance, the maximum and minimum eigenvalues of Hessian with the L2 (cosine) distance show stronger correlation with reconstruction quality in all image similarity scores than the gradient norm in Table~\ref{tab:corr-low}, as expected from our hypothesis. The absolute values of $\sigma_{S}$ are mostly larger than 0.5 for LAVP with the corresponding attack loss function. In the case of cosine distance, LAVP achieves even the optimal value around 0.8. 

In Figure~\ref{fig:corr-low}, values of proxy candidates for each reconstructed sample are plotted in log scale along with its image quality in MSE for CIFAR-10. The gradient norm shows mixed trend in terms of correlation sign as it shows a slightly upward-sloping distribution with $\sigma_{S} = 0.35$ in Figure~\ref{fig:mse-gn-l2-cifar10} but a slightly downward-sloping distribution with $\sigma_{S} = -0.28$ in~\ref{fig:mse-gn-sim-cifar10}. In contrast, LAVP consistently shows upward-sloping distributions with at most $\sigma_{S} = 0.64$ which corresponds to stronger correlation than the gradient norm in Figures~\ref{fig:mse-max-l2-cifar10},~\ref{fig:mse-min-l2-cifar10},~\ref{fig:mse-angmax-sim-cifar10}, and~\ref{fig:mse-angmin-sim-cifar10}.

In Figure~\ref{fig:corr-high}, candidate proxy values for each reconstructed sample are plotted in log scale along with its image quality in MSE on different loss functions and architectures for ImageNet. In Figure~\ref{fig:mse-gn-l2-imagenet}, the gradient norm shows the moderate upward-sloping distribution with $\sigma_{S} = 0.66$, but this phenomenon rather negates the previous hypothesis that examples with higher gradient norm are more vulnerable. Therefore, we believe that this moderate correlation in the case of L2 distance  might be due to the gradient scale, which affects both the gradient norm and Jacobian. In Figure~\ref{fig:mse-gn-sim-imagenet}, the gradient norm shows almost no correlation with $\sigma_{S} = -0.06$ for cosine distance. For the case of cosine, there is no gradient scale issue since a normalizing factor $\frac{1}{||g^{*}||^2}$ exists in Equation~\ref{hes-cos}. In Figures~\ref{fig:mse-max-l2-imagenet},~\ref{fig:mse-min-l2-imagenet},~\ref{fig:mse-angmax-sim-imagenet}, and~\ref{fig:mse-angmin-sim-imagenet}, LAVP consistently shows upward-sloping distributions with at most $\sigma_{S} = 0.74$.

\subsection{LAVP Fusion for Black-box Scenario}
\label{results:cr2}
In a black-box scenario where clients lack knowledge of the attacker's loss function, LAVP should be computed for each potential candidate loss function. To mitigate this complexity, we suggest a loss-agnostic version as a fusion of LAVPs for L2 and cosine distances. In Table~\ref{tab:loss-agnostic}, we present a specific instance of this fusion as the geometric mean between the maximum eigenvalue of the Hessian for L2 loss and the minimum eigenvalue for cosine similarity loss. For both L2 and cosine distances, the loss-agnostic LAVP shows stronger $\sigma_{S}$ than the gradient norm with the vulnerability in most cases. The efficacy of loss-agnostic LAVP can be attributed to the observed minimal correlation between LAVPs for different loss functions. In Table~\ref{tab:corr-high}, LAVP tailored for L2 distance exhibits the correlation of at most $|\sigma_{S}| = 0.06$ with the quality of reconstructed from cosine distance in MSE. On the other hand, LAVP tailored for cosine distance exhibits the correlation of at most $|\sigma_{S}| = 0.07$ with the quality of reconstructed images from L2 distance in MSE. This mutual lack of correlation underlines the absence of any interfering effect between the LAVPs designed for L2 and cosine distances. The concept of LAVP fusion can be extended to any future loss function for gradient matching beyond L2 and cosine.

\section{Conclusion}
\label{sec:conclusion}
This paper introduces a novel concept: a loss-aware vulnerability proxy, called LAVP, designed to gauge the loss-specific quality of reconstructed input from gradient inversion attacks. Unlike the gradient norm, a common heuristic in prior studies, LAVP—represented by either the maximum or minimum eigenvalue of Hessian with respect to gradient matching at ground truth—can explain different reconstruction patterns corresponding to different loss functions for gradient matching in the attack.

This innovation is based on our theoretical results concerning gradient matching optimization. In our theorems, we claim that low bi-Lipschitz constants of the gradient function with respect to the input signify susceptibility to gradient inversion attacks. We also establish a connection between bi-Lipschitz constants of the gradient function and the maximum and minimum eigenvalues of the Hessian near the ground-truth, which is how LAVP is derived.

In our experiments, we show the efficacy of our approach across diverse architectures and datasets, even encompassing high-resolution images like ImageNet. The results indicate that LAVP offers a stronger correlation with vulnerability compared to the gradient norm. We also introduce a loss-agnostic fusion of LAVPs for L2 and cosine distances as the proxy that caters to both L2 and cosine at once. This study not only highlights the significance of Hessian eigenvalues as proxies for vulnerability in gradient inversion attacks but also provides deeper insights into the mechanics of these attacks, paving the way for future research in this domain.

\section{Acknowledgments}
This work was conducted by Center for Applied Research in Artificial Intelligence(CARAI) grant funded by Defense Acquisition Program Administration(DAPA) and Agency for Defense Development(ADD) (UD230017TD). This work was also supported by Institute of Information \& communications Technology Planning \& Evaluation (IITP) grant funded by the Korea government(MSIT) 
(No.2019-0-00075, Artificial Intelligence Graduate School Program(KAIST)).

\bigskip
\noindent Special thanks to Seunghee Koh for thoughtful discussions about the presentation of this work.

\bibliography{aaai24}

\begin{thebibliography}{29}
\providecommand{\natexlab}[1]{#1}

\bibitem[{Dang et~al.(2021)Dang, Thakkar, Ramaswamy, Mathews, Chin, and Beaufays}]{dang2021revealing}
Dang, T.; Thakkar, O.; Ramaswamy, S.; Mathews, R.; Chin, P.; and Beaufays, F. 2021.
\newblock Revealing and Protecting Labels in Distributed Training.
\newblock \emph{Advances in Neural Information Processing Systems}, 34.

\bibitem[{Deng et~al.(2009)Deng, Dong, Socher, Li, Li, and Fei-Fei}]{deng2009imagenet}
Deng, J.; Dong, W.; Socher, R.; Li, L.-J.; Li, K.; and Fei-Fei, L. 2009.
\newblock Imagenet: A large-scale hierarchical image database.
\newblock In \emph{2009 IEEE conference on computer vision and pattern recognition}, 248--255. Ieee.

\bibitem[{Geiping et~al.(2020)Geiping, Bauermeister, Dr{\"o}ge, and Moeller}]{geiping2020inverting}
Geiping, J.; Bauermeister, H.; Dr{\"o}ge, H.; and Moeller, M. 2020.
\newblock Inverting gradients-how easy is it to break privacy in federated learning?
\newblock \emph{Advances in Neural Information Processing Systems}, 33: 16937--16947.

\bibitem[{Hatamizadeh et~al.(2022)Hatamizadeh, Yin, Molchanov, Myronenko, Li, Dogra, Feng, Flores, Kautz, Xu et~al.}]{hatamizadeh2022gradient}
Hatamizadeh, A.; Yin, H.; Molchanov, P.; Myronenko, A.; Li, W.; Dogra, P.; Feng, A.; Flores, M.~G.; Kautz, J.; Xu, D.; et~al. 2022.
\newblock Do Gradient Inversion Attacks Make Federated Learning Unsafe?
\newblock \emph{arXiv preprint arXiv:2202.06924}.

\bibitem[{He et~al.(2016)He, Zhang, Ren, and Sun}]{he2016deep}
He, K.; Zhang, X.; Ren, S.; and Sun, J. 2016.
\newblock Deep residual learning for image recognition.
\newblock In \emph{Proceedings of the IEEE conference on computer vision and pattern recognition}, 770--778.

\bibitem[{Howard()}]{imagewang}
Howard, J. ????
\newblock Imagewang.

\bibitem[{Huang et~al.(2021)Huang, Gupta, Song, Li, and Arora}]{huang2021evaluating}
Huang, Y.; Gupta, S.; Song, Z.; Li, K.; and Arora, S. 2021.
\newblock Evaluating gradient inversion attacks and defenses in federated learning.
\newblock \emph{Advances in Neural Information Processing Systems}, 34.

\bibitem[{Jeon et~al.(2021)Jeon, Lee, Oh, Ok et~al.}]{jeon2021gradient}
Jeon, J.; Lee, K.; Oh, S.; Ok, J.; et~al. 2021.
\newblock Gradient inversion with generative image prior.
\newblock \emph{Advances in Neural Information Processing Systems}, 34: 29898--29908.

\bibitem[{Kariyappa et~al.(2023)Kariyappa, Guo, Maeng, Xiong, Suh, Qureshi, and Lee}]{kariyappa2023cocktail}
Kariyappa, S.; Guo, C.; Maeng, K.; Xiong, W.; Suh, G.~E.; Qureshi, M.~K.; and Lee, H.-H.~S. 2023.
\newblock Cocktail party attack: Breaking aggregation-based privacy in federated learning using independent component analysis.
\newblock In \emph{International Conference on Machine Learning}, 15884--15899. PMLR.

\bibitem[{Kingma and Ba(2015)}]{kingma2015adam}
Kingma, D.~P.; and Ba, J. 2015.
\newblock Adam: A Method for Stochastic Optimization.
\newblock In \emph{ICLR (Poster)}.

\bibitem[{Kone{\v{c}}n{\`y} et~al.(2016)Kone{\v{c}}n{\`y}, McMahan, Yu, Richt{\'a}rik, Suresh, and Bacon}]{konevcny2016federated}
Kone{\v{c}}n{\`y}, J.; McMahan, H.~B.; Yu, F.~X.; Richt{\'a}rik, P.; Suresh, A.~T.; and Bacon, D. 2016.
\newblock Federated learning: Strategies for improving communication efficiency.
\newblock \emph{arXiv preprint arXiv:1610.05492}.

\bibitem[{Krizhevsky(2009)}]{alex2009learning}
Krizhevsky, A. 2009.
\newblock Learning Multiple Layers of Features From Tiny Images.

\bibitem[{Krizhevsky, Sutskever, and Hinton(2017)}]{krizhevsky2017imagenet}
Krizhevsky, A.; Sutskever, I.; and Hinton, G.~E. 2017.
\newblock Imagenet classification with deep convolutional neural networks.
\newblock \emph{Communications of the ACM}, 60(6): 84--90.

\bibitem[{LeCun et~al.(1998)LeCun, Bottou, Bengio, and Haffner}]{lecun1998gradient}
LeCun, Y.; Bottou, L.; Bengio, Y.; and Haffner, P. 1998.
\newblock Gradient-based learning applied to document recognition.
\newblock \emph{Proceedings of the IEEE}, 86(11): 2278--2324.

\bibitem[{Li et~al.(2022)Li, Sun, Yang, Gao, Zhang, Xie, Smith, and Wang}]{li2021label}
Li, O.; Sun, J.; Yang, X.; Gao, W.; Zhang, H.; Xie, J.; Smith, V.; and Wang, C. 2022.
\newblock Label leakage and protection in two-party split learning.

\bibitem[{Ma et~al.(2022)Ma, Sun, Cui, Li, Guan, and Liu}]{ma2022instance}
Ma, K.; Sun, Y.; Cui, J.; Li, D.; Guan, Z.; and Liu, J. 2022.
\newblock Instance-wise Batch Label Restoration via Gradients in Federated Learning.
\newblock In \emph{The Eleventh International Conference on Learning Representations}.

\bibitem[{Ma et~al.(2018)Ma, Zhang, Zheng, and Sun}]{ma2018shufflenet}
Ma, N.; Zhang, X.; Zheng, H.-T.; and Sun, J. 2018.
\newblock Shufflenet v2: Practical guidelines for efficient cnn architecture design.
\newblock In \emph{Proceedings of the European conference on computer vision (ECCV)}, 116--131.

\bibitem[{McMahan et~al.(2017)McMahan, Moore, Ramage, Hampson, and y~Arcas}]{mcmahan2017communication}
McMahan, B.; Moore, E.; Ramage, D.; Hampson, S.; and y~Arcas, B.~A. 2017.
\newblock Communication-efficient learning of deep networks from decentralized data.
\newblock In \emph{Artificial intelligence and statistics}, 1273--1282. PMLR.

\bibitem[{Mo et~al.(2021)Mo, Borovykh, Malekzadeh, Demetriou, G{\"u}nd{\"u}z, and Haddadi}]{moquantifying}
Mo, F.; Borovykh, A.; Malekzadeh, M.; Demetriou, S.; G{\"u}nd{\"u}z, D.; and Haddadi, H. 2021.
\newblock Quantifying and Localizing Usable Information Leakage from Neural Network Gradients.
\newblock \emph{European Symposium on Research in Computer Security}.

\bibitem[{Simonyan and Zisserman(2014)}]{simonyan2014very}
Simonyan, K.; and Zisserman, A. 2014.
\newblock Very deep convolutional networks for large-scale image recognition.
\newblock \emph{arXiv preprint arXiv:1409.1556}.

\bibitem[{Wainakh et~al.(2021)Wainakh, M{\"u}{\ss}ig, Grube, and M{\"u}hlh{\"a}user}]{wainakh2021label}
Wainakh, A.; M{\"u}{\ss}ig, T.; Grube, T.; and M{\"u}hlh{\"a}user, M. 2021.
\newblock Label leakage from gradients in distributed machine learning.
\newblock In \emph{2021 IEEE 18th Annual Consumer Communications \& Networking Conference (CCNC)}, 1--4. IEEE.

\bibitem[{Wang et~al.(2004)Wang, Bovik, Sheikh, and Simoncelli}]{wang2004image}
Wang, Z.; Bovik, A.~C.; Sheikh, H.~R.; and Simoncelli, E.~P. 2004.
\newblock Image quality assessment: from error measurement to structural similarity.
\newblock \emph{IEEE transactions on image processing}, 13(1).

\bibitem[{Ye et~al.(2022)Ye, Zhu, Zhou, Liu, and Zhou}]{ye2022label}
Ye, D.; Zhu, T.; Zhou, S.; Liu, B.; and Zhou, W. 2022.
\newblock Label-only Model Inversion Attack: The Attack that Requires the Least Information.
\newblock \emph{arXiv preprint arXiv:2203.06555}.

\bibitem[{Yin et~al.(2021)Yin, Mallya, Vahdat, Alvarez, Kautz, and Molchanov}]{yin2021see}
Yin, H.; Mallya, A.; Vahdat, A.; Alvarez, J.~M.; Kautz, J.; and Molchanov, P. 2021.
\newblock See through gradients: Image batch recovery via gradinversion.
\newblock In \emph{Proceedings of the IEEE/CVF Conference on Computer Vision and Pattern Recognition}, 16337--16346.

\bibitem[{Zhang et~al.(2018)Zhang, Isola, Efros, Shechtman, and Wang}]{zhang2018unreasonable}
Zhang, R.; Isola, P.; Efros, A.~A.; Shechtman, E.; and Wang, O. 2018.
\newblock The unreasonable effectiveness of deep features as a perceptual metric.
\newblock In \emph{Proceedings of the IEEE conference on computer vision and pattern recognition}, 586--595.

\bibitem[{Zhao, Mopuri, and Bilen(2020)}]{zhao2020idlg}
Zhao, B.; Mopuri, K.~R.; and Bilen, H. 2020.
\newblock idlg: Improved deep leakage from gradients.
\newblock \emph{arXiv preprint arXiv:2001.02610}.

\bibitem[{Zhu and Blaschko(2020)}]{zhu2020r}
Zhu, J.; and Blaschko, M.~B. 2020.
\newblock R-GAP: Recursive Gradient Attack on Privacy.
\newblock In \emph{International Conference on Learning Representations}.

\bibitem[{Zhu, Yao, and Blaschko(2023)}]{zhu2023surrogate}
Zhu, J.; Yao, R.; and Blaschko, M.~B. 2023.
\newblock Surrogate model extension (SME): A fast and accurate weight update attack on federated learning.

\bibitem[{Zhu, Liu, and Han(2019)}]{zhu2019deep}
Zhu, L.; Liu, Z.; and Han, S. 2019.
\newblock Deep leakage from gradients.
\newblock \emph{Advances in Neural Information Processing Systems}, 32.

\end{thebibliography}

\newpage
\pagebreak
\noindent\textbf{\Large Appendix}
\vspace{1em}

We include additional information in the Appendix.

We provide mathematical proofs for several theorems in the main paper.

We provide the pseudocode for computing maximum eigenvalues of hessian, which is omitted in the main paper due to page limits.

We present limitations and future work section, which is omitted in the main paper due to space constraints.

We present qualitative comparison between gradient norm and LAVP (ours) on datasets including CIFAR100 (Figure 4), ImageNette (Figure 5), and ImageWoof (Figure 6).

We present how strong LAVP correlates to final loss in local optimization scenario to understand the effectiveness of LAVP in capturing optimization behavior.

\section{Mathematical Proofs}
\label{sec:proof}

\newtheorem*{theorem*}{Theorem}

\subsection{A1. Proof of Theorem 1}
\begin{theorem*}
\normalfont (The first gradient inversion loss theorem). Suppose $g_w(x)$ is Lipschitz continuous with respect to $x$ with constant $L$ and $\mathcal{L}_{grad}(x) = ||g_w(x) - g^{*}||_2^2 $ is the gradient matching loss. Then, when gradient descent $\Delta x$ is applied with step size $\mu = \frac{1}{2L^2} > 0$ and $L > \epsilon$ for some $\epsilon > 0$, the following holds:

\begin{equation*}
    \mathcal{L}_{grad}(x+ \Delta x) \leq \mathcal{L}_{grad}(x) - \frac{1}{4L^2} ||\frac{\partial \mathcal{L}_{grad}^{x}}{\partial x}||_{2}^{2},
\end{equation*}
where $L > \epsilon$ satisfies $|| \mu \Delta x|| < \delta$ such that $g_w(x + \mu \Delta x) = g_w(x) + \mu \nabla_x g_w(x) \Delta x$ holds approximately. 
\label{monotonic}
\end{theorem*}

\begin{proof}

First, we will compute the vector for gradient descent, $\Delta x = -\mu \frac{\partial \mathcal{L}_{grad}}{\partial x}$ by chain rule as follows:

\begin{align*}
\Delta x 
&= -\mu \frac{\partial \mathcal{L}_{grad}}{\partial x} \\
&= -\mu \frac{\partial ||\nabla_w \mathcal{L}(f(x), y) - g^{*}||_2^2}{\partial x}  \\
&= -2\mu \nabla_{x} \nabla_{w} \mathcal{L}(f(x),y) (\nabla_w \mathcal{L}(f(x),y) - g^{*}).
\end{align*}

Then, $\mathcal{L}_{grad}(x+ \Delta x)$ can be separated into three terms by summation like the following:

\begin{align*}
\mathcal{L}_{grad}(x+ \Delta x) &= ||\nabla_w \mathcal{L}(f(x + \Delta x), y) - g^{*}||_2^2 \\
                                    &= ||\nabla_w \mathcal{L}(f(x + \Delta x), y) - \nabla_w \mathcal{L}(f(x), y) + \\ 
                                    & \quad\quad \nabla_w \mathcal{L}(f(x), y) - g^{*}||_2^2 \\   
                                    &= ||u||_2^2 + 2u^{T}v + ||v||_2^2
\end{align*}
, where $u = \nabla_w \mathcal{L}(f(x + \Delta x), y) - \nabla_w \mathcal{L}(f(x), y)$ and $v = \nabla_w \mathcal{L}(f(x), y) - g^{*}$.\\

For the  first term, $||u||_2^2 = ||\nabla_w \mathcal{L}(f(x + \Delta x), y) - \nabla_w \mathcal{L}(f(x), y)||_2^2 \leq L^2||\Delta x||^2 = L^2\mu^2 ||\frac{\partial \mathcal{L}_{grad}}{\partial x}||_2^2 $ due to the $L$-Lipschitz continuity condition of $\nabla_w \mathcal{L}(f(x), y)$ with respect to an input.\\

For the second term, 
\begin{align*}
2u^{T}v &= 2 (\nabla_w \mathcal{L}(f(x + \Delta x), y) - \nabla_w \mathcal{L}(f(x), y))^{T}\\
&\quad\quad (\nabla_w \mathcal{L}(f(x), y) - g^{*}) \\
 &\approx 2 \Delta x ^{T} \nabla_{x} \nabla_{w} \mathcal{L}(f(x), y) (\nabla_w \mathcal{L}(f(x), y) - g^{*}) \\
 &(\because \text{Taylor's first-order approximation with } \mu < \delta )\\
&= -\mu (\frac{\partial \mathcal{L}_{grad}}{\partial x})^{T}(\frac{\partial \mathcal{L}_{grad}}{\partial x})\\
&= -\mu ||\frac{\partial \mathcal{L}_{grad}}{\partial x}||_2^2 \\
&(\because \text{Recall that how $\Delta x$ is computed in the first step} ). \\
\end{align*}

For the third term, $||v||_2^2 = ||\nabla_w \mathcal{L}(f(x), y) - g^{*}||_2^2 = \mathcal{L}_{grad}(x) $. \\

Then, summing up the three terms considered above will lead to the following inequality: \\

\begin{align*}
\mathcal{L}_{grad}(x+ \Delta x) &\leq \mathcal{L}_{grad}(x) + (L^2\mu^2 - \mu) ||\frac{\partial \mathcal{L}_{grad}}{\partial x}||_2^2 \\
&= \mathcal{L}_{grad}(x) - \frac{1}{4L^2} ||\frac{\partial \mathcal{L}_{grad}}{\partial x}||_2^2  \\
& \quad (\because \text{min. at } \mu = \frac{1}{2L^2}).
\end{align*}

For both $\mu = \frac{1}{2L^2}$ and $\mu < \delta$ to be met, $\frac{1}{2L^2} < \delta$ should be satisfied and this is why the premise that $L > \epsilon = \frac{1}{\sqrt{2\delta}}$ is required for the theorem. We empirically found that $\tilde{L}$, the estimated value for $L$, mostly turned out to be not small, thus meeting the premise in practice. We will derive the theorem for small $L$ to explain outliers in future work. 

\end{proof}

\subsection{A2. Proof of Theorem 2}

\begin{theorem*}
\normalfont (The second gradient inversion loss theorem). Suppose $|| g_w(x_1) - g_w(x_2)|| \geq M || x_1 - x_2|| \forall x_1, x_2$ for $M > 0$ and $\mathcal{L}_{grad}(x) = ||g_w(x) - g^{*}||_2^2 $ is the gradient matching loss. Then, when gradient descent $\Delta x$ is applied with step size $\mu  < \delta_1$ for some $\delta_1 > 0$, the following holds:

\begin{equation*}
    \mathcal{L}_{grad}(x+ \Delta x) \geq \mathcal{L}_{grad}(x) - \frac{1}{4M^2} ||\frac{\partial \mathcal{L}_{grad}^{x}}{\partial x}||_{2}^{2},
\end{equation*}
$\mu < \delta_1$ satisfies $|| \mu \Delta x|| < \delta$ such that $g_w (x + \mu \Delta x) = g_w (x) +  \mu \nabla_x g_w(x) \Delta x$ holds approximately.
\label{monotonic2}
\end{theorem*}

\begin{proof}
Similar to the proof of previous theorem, we can use the following equation again:
\begin{align*}
\mathcal{L}_{grad}(x+ \Delta x) &= ||\nabla_w \mathcal{L}(f(x + \Delta x), y) - g^{*}||_2^2 \\
                                    &= ||\nabla_w \mathcal{L}(f(x + \Delta x), y) - \nabla_w \mathcal{L}(f(x), y) + \\ 
                                    & \quad\quad \nabla_w \mathcal{L}(f(x), y) - g^{*}||_2^2 \\   
                                    &= ||u||_2^2 + 2u^{T}v + ||v||_2^2
\end{align*}
, where $u = \nabla_w \mathcal{L}(f(x + \Delta x), y) - \nabla_w \mathcal{L}(f(x), y)$ and $v = \nabla_w \mathcal{L}(f(x), y) - g^{*}$.\\

For the  first term, $||u||_2^2 = ||\nabla_w \mathcal{L}(f(x + \Delta x), y) - \nabla_w \mathcal{L}(f(x), y)||_2^2 \geq M^2||\Delta x||^2 = M^2\mu^2 ||\frac{\partial \mathcal{L}_{grad}}{\partial x}||_2^2 $ due to the given condition.\\

For the second term,
\begin{align*}
2u^{T}v &= 2 (\nabla_w \mathcal{L}(f(x + \Delta x), y) - \nabla_w \mathcal{L}(f(x), y))^{T}\\
&\quad\quad (\nabla_w \mathcal{L}(f(x), y) - g^{*}) \\
 &\approx 2 \Delta x ^{T} \nabla_{x} \nabla_{w} \mathcal{L}(f(x), y) (\nabla_w \mathcal{L}(f(x), y) - g^{*}) \\
 &(\because \text{Taylor's first-order approximation with } \mu < \delta ) \\
&= -\mu (\frac{\partial \mathcal{L}_{grad}}{\partial x})^{T}(\frac{\partial \mathcal{L}_{grad}}{\partial x}) \\
&= -\mu ||\frac{\partial \mathcal{L}_{grad}}{\partial x}||_2^2 \\
&(\because \text{Recall that how $\Delta x$ is computed in the first step} ). \\
\end{align*}

For the third term, $||v||_2^2 = ||\nabla_w \mathcal{L}(f(x), y) - g^{*}||_2^2 = \mathcal{L}_{grad}(x) $. \\

Then, summing up the three terms considered above will lead to the following inequality: \\

\begin{align*}
\mathcal{L}_{grad}(x+ \Delta x) &\geq \mathcal{L}_{grad}(x) + (M^2\mu^2 - \mu) ||\frac{\partial \mathcal{L}_{grad}}{\partial x}||_2^2 \\
&\geq \mathcal{L}_{grad}(x) - \frac{1}{4M^2} ||\frac{\partial \mathcal{L}_{grad}}{\partial x}||_2^2 \\
&(\because \text{min. at } \mu = \frac{1}{2M^2}).
\end{align*}

Unlike the case for Theorem 1, the inequality above holds for any $\mu$, thus no restriction on $\mu$.
\end{proof}

\subsection{A3. Proof of Theorem 3}
\begin{theorem*}
\normalfont (Hessian at ground truth for L2 distance). Suppose $\mathcal{L}_{grad}$ is L2 distance, then the hessian at ground truth is like the following:

\begin{equation*}
    H_{\text{L2}}(x^{*}) = J(x^{*})^{T}J(x^{*}),
\end{equation*}
where $J(x^{*})$ is the Jacobian of gradient with respect to input variable (i.e., $J(x^{*}) = \nabla_{x=x^{*}}g_w(x)$).
\label{hessian-l2}
\end{theorem*}

\begin{proof}
Note that $\mathcal{L}_{grad}(x) = \frac{1}{2}||\nabla_w \mathcal{L}(f(x), y) - g^{*}||_2^2$. Then, $\frac{\partial \mathcal{L}_{grad}(x)}{\partial x} = J(x)^{T}(\nabla_w \mathcal{L}(f(x), y) - g^{*})$. Then, the hessian would be $\frac{\partial}{\partial x}\frac{\partial L_{grad}(x)}{\partial x} = \frac{\partial J(x)^{T}}{\partial x}(\nabla_w \mathcal{L}(f(x), y) - g^{*}) + J(x)^{T}J(x)$ (by Product Rule). Note that $\nabla_w \mathcal{L}(f(x^{*}), y) = g^{*} $. Thus, replacing $x$ with $x^{*}$ cancels the former term, thus $H_{\text{L2}}(x^{*}) = \frac{\partial}{\partial x}\frac{\partial \mathcal{L}_{grad}(x)}{\partial x}|_{x = x^{*}} = J(x^{*})^{T}J(x^{*})$ holds.  
\end{proof}

\subsection{A4. Proof of Theorem 4}
\begin{theorem*}
\normalfont (Hessian at ground truth for cosine distance). Suppose $\mathcal{L}_{grad}$ is cosine distance, then the hessian at ground truth is like the following:

\begin{equation*}
    H_{\text{cos}}(x^{*}) = \frac{1}{||g^{*}||_2^2}J(x^{*})^{T}(I - \frac{g^{*}}{||g^{*}||_2}\frac{g^{*T}}{||g^{*}||_2})J(x^{*}),
\end{equation*}
$I$ is the identity matrix.
\label{hessian-cos}
\end{theorem*}

\begin{proof}
Note that $\mathcal{L}_{grad}(x) = 1 - \frac{g^{*}}{||g^{*}||_2^2}^{T}\frac{\nabla_w \mathcal{L}(f(x), y)}{||\nabla_w \mathcal{L}(f(x), y)||_2}$.

Let $v$ denote $ \frac{\nabla_w \mathcal{L}(f(x), y)}{||\nabla_w \mathcal{L}(f(x), y)||_2}$ and $h$ denote $\nabla_w \mathcal{L}(f(x), y)$. 

Then, $\frac{\partial L_{grad}(x)}{\partial x} =\frac{\partial h}{\partial x} \frac{\partial v}{\partial h} \frac{\partial h}{\partial v}   $ (by Chain Rule) = $- J(x)^{T}\frac{1}{||h||} (I - \frac{1}{||h||^2}h h^{T})\frac{g^{*}}{||g^{*}||_2}$. 

Then, hessian $\frac{\partial}{\partial x}\frac{\partial L_{grad}(x)}{\partial x} = \frac{\partial}{\partial x} (-J(x)^{T}\frac{1}{||h||_2})(I - \frac{1}{||h||_2^2}h h^{T})\frac{g^{*}}{||g^{*}||_2} -J(x)^{T}\frac{1}{||h||_2} \frac{\partial}{\partial x} (I - \frac{1}{||h||_2^2}h h^{T})\frac{g^{*}}{||g^{*}||_2}$. 

By replacing $x$ with $x^{*}$, the former term is canceled out because $(I - \frac{1}{||h||_2^2}h h^{T})\frac{g^{*}}{||g^{*}||_2} = (I - \frac{1}{||g^{*}||_2^2}g^{*}g^{*T})\frac{g^{*}}{||g^{*}||_2} = 0$. 

Then, the latter term is only left, thus hessian would be $-J(x)^{T}\frac{1}{||h||_2} \frac{\partial}{\partial x} (I - \frac{1}{||h||_2^2}h h^{T})\frac{g^{*}}{||g^{*}||_2} =2J(x)^{T}\frac{1}{||h||_2^2} \frac{\partial}{\partial x} (I - \frac{1}{||h||_2^2}h h^{T})J(x) \frac{h}{||h||_2}^{T}\frac{g^{*}}{||g^{*}||_2}$. Using the substitution $x = x^{*}$ (then, $h = g^{*}$), hessian would be like the following: $2J(x^{*})^{T}\frac{1}{||g^{*}||_2^2} \frac{\partial}{\partial x} (I - \frac{1}{||g^{*}||_2^2}g^{*} g^{*T})J(x^{*}) \frac{g^{*}}{||g^{*}||_2}^{T}\frac{g^{*}}{||g^{*}||_2} = 2J(x^{*})^{T}\frac{1}{||g^{*}||_2^2} \frac{\partial}{\partial x} (I - \frac{1}{||g^{*}||_2^2}g^{*} g^{*T})J(x^{*})$, thus the theorem holds. 
\end{proof}

\subsection{A5. The correlation between $L$ and $M$ (key variables in section 4.1) to the maximum and the minimum eigenvalues of the Hessian.}
By Taylor's second-order approximation, $\mathcal{L}_{grad}(x^{*} + \Delta x) = \mathcal{L}_{grad}(x^{*}) + \Delta x^{T} \nabla_{x=x^{*}} \mathcal{L}_{grad}(x) + \frac{1}{2} \Delta x^{T} H \left(x^{*} \right) \Delta x$. For $\mathcal{L}_{grad}$ being either L2 or cosine distance, $\mathcal{L}_{grad}(x^{*}) = 0$ and  $\nabla_{x=x^{*}} \mathcal{L}_{grad}(x) = \mathbf{0}$ hold (see the proofs in A3. and A4.).
Then, we can rewrite $ \mathcal{L}_{grad}(x^{*} + \Delta x) = \frac{1}{2} \Delta x^{T} H \left(x^{*} \right) \Delta x$ and the following holds.
\begin{align*}
 &\frac{ \left| \left| \nabla_{w} \mathcal{L}(f(x^{*} + \Delta x), y) -  \nabla_{w} \mathcal{L}(f(x^{*}), y) \right | \right |_{2}}{\left | \left | \Delta x \right | \right |_{2}} \\
 &= \frac{\sqrt{\mathcal{L}_{grad}(x^{*} + \Delta x) } }{ \left | \left | \Delta x \right | \right |_{2}}\\
 &= \frac{\sqrt{\mathcal{L}_{grad}(x^{*} + \Delta x) - \mathcal{L}_{grad}(x^{*})} }{ \left | \left | \Delta x \right | \right |_{2}}  \left( \because \mathcal{L}_{grad}(x^{*}) = 0 \right) \\
 &= \frac{\sqrt{\frac{1}{2} \Delta x^{T} H \left(x^{*} \right) \Delta x } }{ \left | \left | \Delta x \right | \right |_{2}} \\, \text{and}
\end{align*}

$\frac{\sqrt{\lambda_{min}}}{\sqrt{2}} \leq \frac{\sqrt{\frac{1}{2} \Delta x^{T} H \left(x^{*} \right) \Delta x } }{ \left | \left | \Delta x \right | \right |_{2}} \leq \frac{\sqrt{\lambda_{max}}}{\sqrt{2}}, 
$
where $\lambda_{min}$ and $\lambda_{max}$ are minimum and maximum eigenvalues of $H(x^{*})$. Therefore, $\frac{\sqrt{\lambda_{max}}}{\sqrt{2}}$ and $\frac{\sqrt{\lambda_{min}}}{\sqrt{2}}$ provides the lower and upper bounds of bi-Lipschitz constants $L$ and $M$ in Theorem 1 and 2 respectively. Note that $\frac{\sqrt{\lambda_{max}}}{\sqrt{2}}$ and $\frac{\sqrt{\lambda_{min}}}{\sqrt{2}}$ are exactly $L$ and $M$ respectively for the special case when $x_2 = x^{*}$, which is of our interest.

\section{Algorithms}
\label{sec:algo}

In Algorithm~\ref{alg:max_eigenvalue}, we describe how the maximum eigenvalue of Hessian is computed using the pseudocode.

\begin{algorithm}[h]
\caption{Pseudocode for computing maximum eigenvalue of Hessian, PyTorch-like}
\definecolor{codeblue}{rgb}{0.25,0.5,0.5}
\definecolor{codekw}{rgb}{0.85, 0.18, 0.50}
\lstset{
  backgroundcolor=\color{white},
  basicstyle=\fontsize{7.2pt}{7.2pt}\ttfamily\selectfont,
  columns=fullflexible,
  breaklines=true,
  captionpos=b,
  commentstyle=\fontsize{3.5pt}{3.5pt}\color{codeblue},
  keywordstyle=\fontsize{7.5pt}{7.5pt}\color{codekw},
}
\begin{lstlisting}[language=python]
def max_eigenvalue(x, label, model, g_gt):
    # x: ground-truth image
    # label : x's class
    # model : FL model
    # g_gt : model gradient from ground truth

    m_e = 0 # max eigenvalue candidate
    
    for _ in range(N): # N is large enough
        v = torch.randn_like(x) # initialize vector
        v = v/torch.norm(v) # normalize vector

        #initialie gradients to zero
        x_tmp = x.copy()
        x_tmp.grad *= 0.0
        model.zero_grad()

        loss1 = loss_fn(model(x_tmp - e*v), label) #compute loss at neighborhood of x, e is small
    
        gradient1 = torch.autograd.grad(loss1, model.parameters(), create_graph=True) #compute gradient 
    
        g1 = torch.cat([gradient1.view(-1, 1).detach() for g in gradient1], dim=0).squeeze() # flatten gradient into 1-D

        model.zero_grad()

        loss2 = loss_fn(model(x_tmp + e*v), label) #compute loss at neighborhood of x at the opposite side, e is small
    
        gradient2 = torch.autograd.grad(loss2, model.parameters(), create_graph=True) #compute gradient 
    
        g2 = torch.cat([gradient2.view(-1, 1).detach() for g in gradient2], dim=0).squeeze() # flatten gradient into 1-D

        
        g_diff = (g2 - g)/(2*e) # compute difference, which is approximately a Jacobian vector product
            
        g_diff = g_diff.detach()

        # For cosine similiarity loss, compute intermeidate terms with g_gt
        
        if loss_type == 'cosine':
            g_diff = g_diff - (g_diff*g_gt).sum(0)*g_gt
          
        model.zero_grad()
        loss = loss_fn(model(x_tmp), label) # compute loss at gt
        gradient = torch.autograd.grad(loss, model.parameters(), create_graph=True) # compute gradient
        g = gorch.cat([gradient.view(-1, 1).detach() for g in gradient], dim=0).squeeze()
        #flatten gradient into 1-D
        ig = torch.autograd.grad(outputs=g, inputs=x, grad_outputs=g_diff) #Hessian vector product
        
        nrm = (ig[0]*v).sum() #maximum eigenvalue candidate

        m_e = max(m_e, nrm) #update maximum eigenvalue
        
        v = ig[0]
return m_e
\end{lstlisting}
\label{alg:max_eigenvalue}
\end{algorithm}

\section{Limitations and Future Work}
\label{sec:limit}
Our work is focused on pure gradient matching loss for fundamental analysis, without batch statistics matching loss. However, state-of-the-art method currently uses batch statistics matching, thus a theoretical approach on optimizing batch statistics matching is required to craft more advanced proxy for the vulnerability under state-of-the-art gradient inversion attacks.

\section{Plots on Other Datasets}
\label{sec:plots}

\begin{figure*}[t]
  
  \begin{subfigure}{.33\textwidth}
  \centering
  \includegraphics[width=1\linewidth]{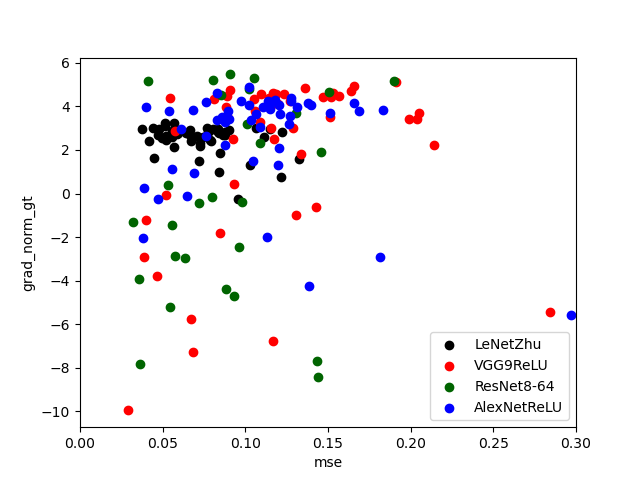}  
  \caption{ grad\_norm vs MSE (L2, $\sigma_{S} = 0.41$)}
  \label{fig:mse-gn-l2-cifar100}
\end{subfigure}
\begin{subfigure}{.33\textwidth}
  \centering
  \includegraphics[width=1\linewidth]{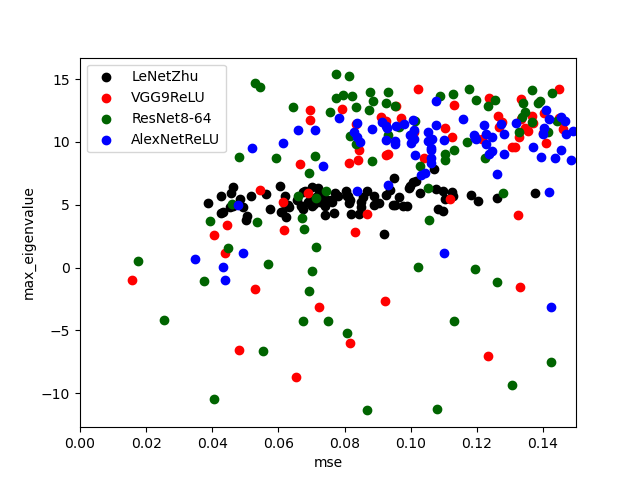}
  \caption{max vs MSE (L2, $\sigma_{S} = 0.46$)}
  \label{fig:mse-max-l2-cifar100}
\end{subfigure}
\begin{subfigure}{.33\textwidth}
  \centering
  \includegraphics[width=1\linewidth]{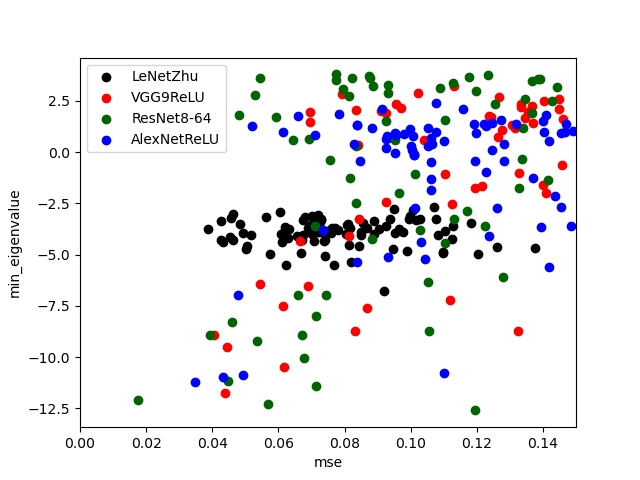}  
  \caption{min vs MSE (L2, $\sigma_{S} = 0.41$)}
  \label{fig:mse-min-l2-cifar100}
\end{subfigure}
\begin{subfigure}{.325\textwidth}
  \centering
  \includegraphics[width=1\linewidth]{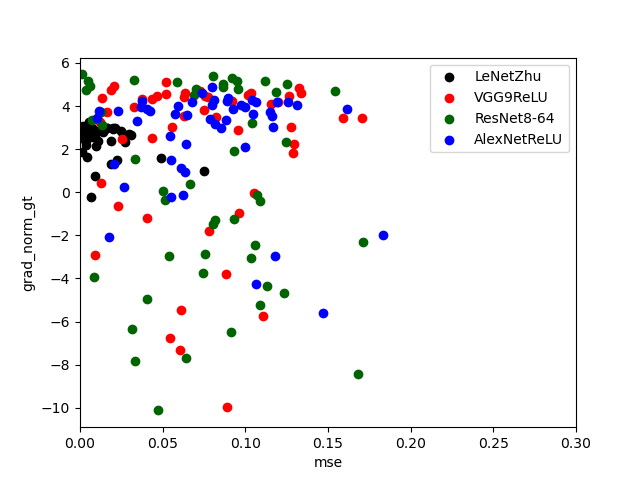}  
  \caption{grad\_norm vs MSE (CS, $\sigma_{S} = 0.1$)}
  \label{fig:mse-gn-sim-cifar100}
\end{subfigure}
\begin{subfigure}{.325\textwidth}
  \centering
  \includegraphics[width=1\linewidth]{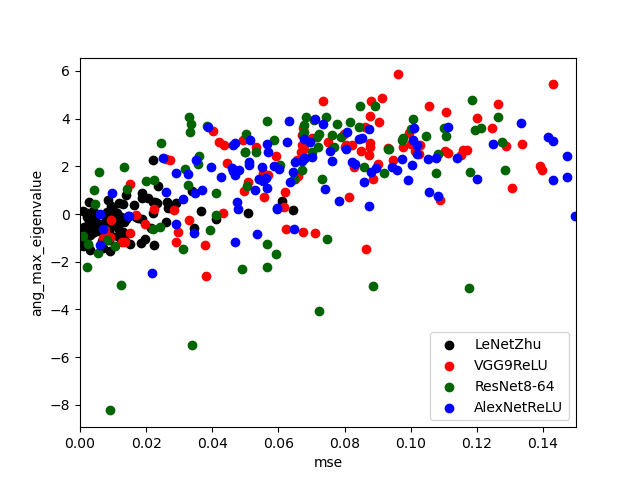}  
  \caption{ang\_max vs MSE (CS, $\sigma_{S} = 0.67$)}
  \label{fig:mse-angmax-cs}
\end{subfigure}
\begin{subfigure}{.325\textwidth}
  \centering
  \includegraphics[width=1\linewidth]{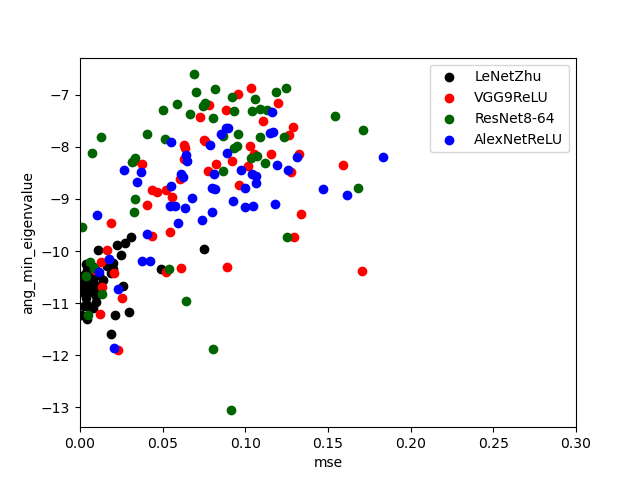}  
  \caption{ang\_min vs MSE (CS,  $\sigma_{S} = 0.69$))}
  \label{fig:mse-angmin-cs}
\end{subfigure}

\caption{\textbf{Comparison of gradient norm, maximum and minimum eigenvalues of Hessian in terms of the correlation with MSE of reconstructed samples over several architectures on CIFAR100 test samples.}}
  
\label{fig:cifar100-corr}
\end{figure*}

\begin{figure*}[t]
  
  \begin{subfigure}{.33\textwidth}
  \centering
  \includegraphics[width=1\linewidth]{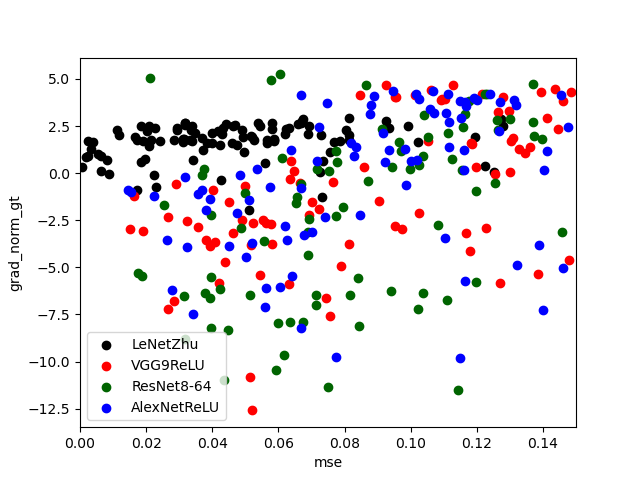}  
  \caption{ grad\_norm vs MSE (L2, $\sigma_{S} = 0.03$)}
  \label{fig:mse-gn-l2-imagenette}
\end{subfigure}
\begin{subfigure}{.33\textwidth}
  \centering
  \includegraphics[width=1\linewidth]{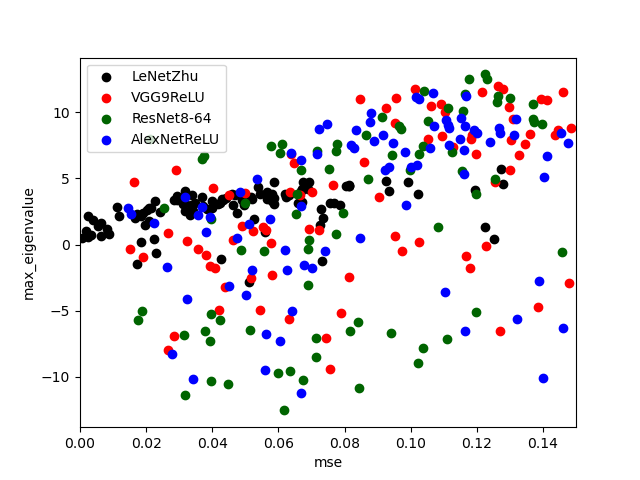}
  \caption{max vs MSE (L2, $\sigma_{S} = 0.33$)}
  \label{fig:mse-max-l2-imagenette}
\end{subfigure}
\begin{subfigure}{.33\textwidth}
  \centering
  \includegraphics[width=1\linewidth]{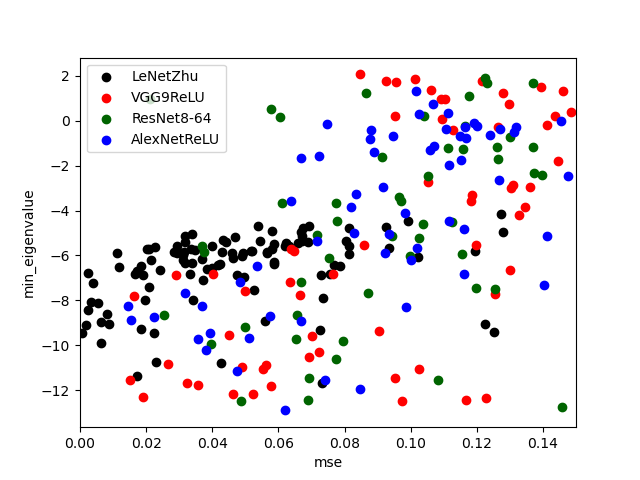}  
  \caption{min vs MSE (L2, $\sigma_{S} = 0.34$)}
  \label{fig:mse-min-l2-imagenette}
\end{subfigure}
\begin{subfigure}{.325\textwidth}
  \centering
  \includegraphics[width=1\linewidth]{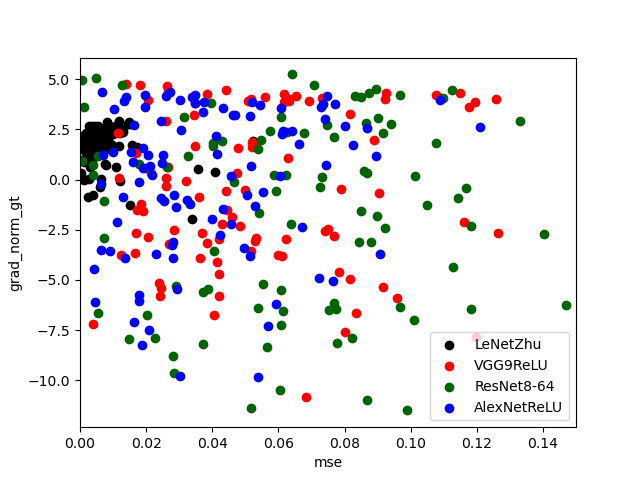}  
  \caption{grad\_norm vs MSE (CS, $\sigma_{S} = -0.13$)}
  \label{fig:mse-gn-sim-imagenette}
\end{subfigure}
\begin{subfigure}{.325\textwidth}
  \centering
  \includegraphics[width=1\linewidth]{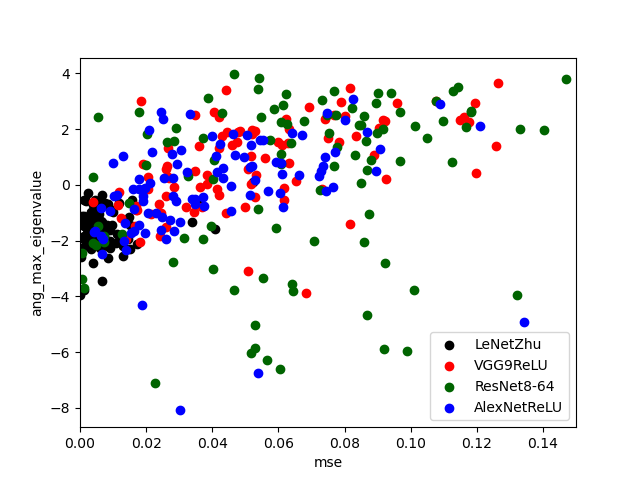}  
  \caption{ang\_max vs MSE (CS, $\sigma_{S} = 0.57$)}
  \label{fig:mse-angmax-cs}
\end{subfigure}
\begin{subfigure}{.325\textwidth}
  \centering
  \includegraphics[width=1\linewidth]{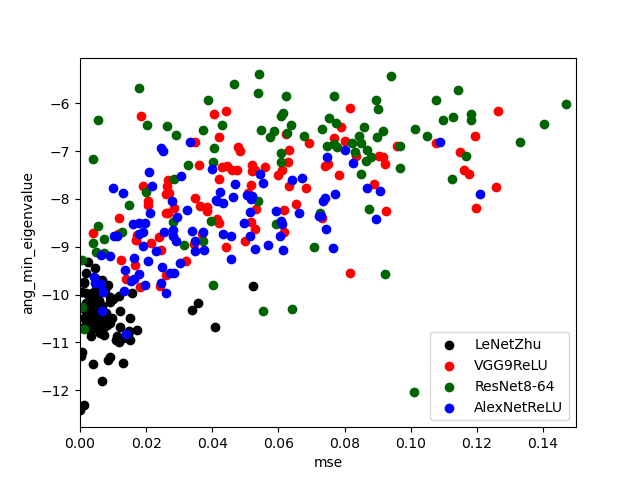}  
  \caption{ang\_min vs MSE (CS,  $\sigma_{S} = 0.75$))}
  \label{fig:corr-imagenette}
\end{subfigure}

\caption{\textbf{Comparison of gradient norm, maximum and minimum eigenvalues of Hessian in terms of the correlation with MSE of reconstructed samples over several architectures on ImageNette test samples.}}
  
\label{fig:imagenette-corr}
\end{figure*}

\begin{figure*}[t]
  
  \begin{subfigure}{.33\textwidth}
  \centering
  \includegraphics[width=1\linewidth]{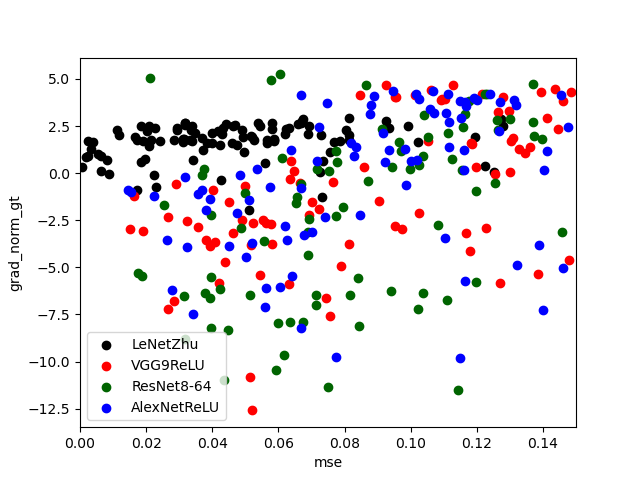}  
  \caption{ grad\_norm vs MSE (L2, $\sigma_{S} = 0.35$)}
  \label{fig:mse-gn-l2-imagewoof}
\end{subfigure}
\begin{subfigure}{.33\textwidth}
  \centering
  \includegraphics[width=1\linewidth]{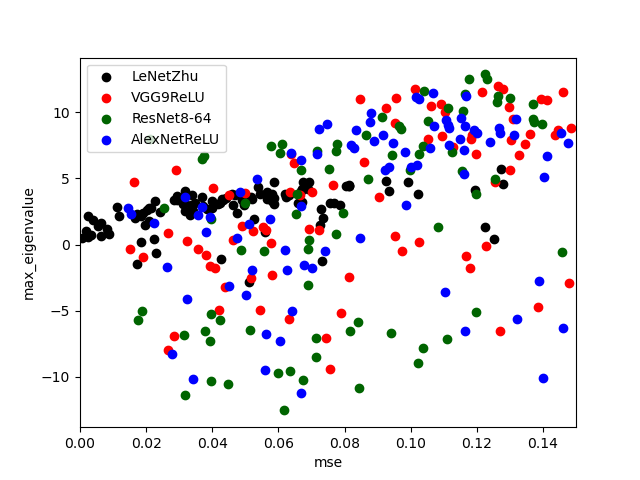}
  \caption{max vs MSE (L2, $\sigma_{S} = 0.66$)}
  \label{fig:mse-max-l2-imagewoof}
\end{subfigure}
\begin{subfigure}{.33\textwidth}
  \centering
  \includegraphics[width=1\linewidth]{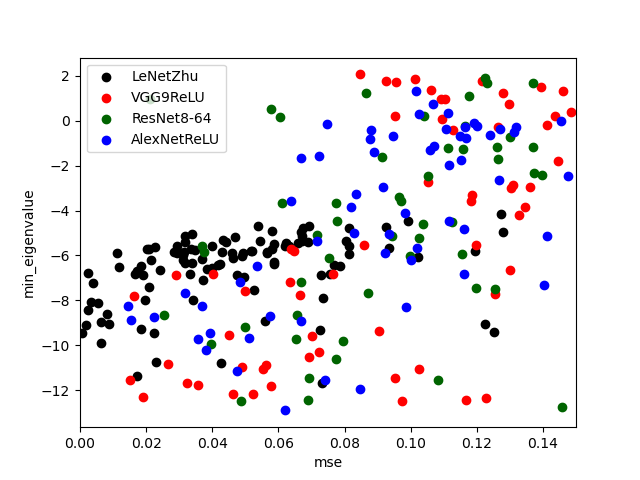}  
  \caption{min vs MSE (L2, $\sigma_{S} = 0.68$)}
  \label{fig:mse-min-l2-imagewoof}
\end{subfigure}
\begin{subfigure}{.325\textwidth}
  \centering
  \includegraphics[width=1\linewidth]{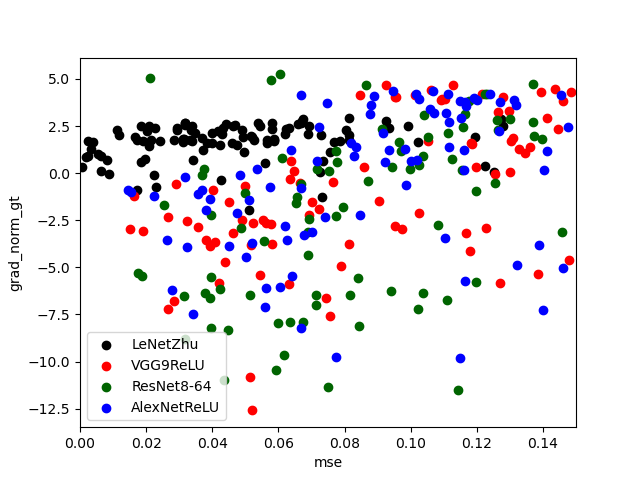}  
  \caption{grad\_norm vs MSE (CS, $\sigma_{S} = 0.00$)}
  \label{fig:mse-gn-sim-imagewoof}
\end{subfigure}
\begin{subfigure}{.325\textwidth}
  \centering
  \includegraphics[width=1\linewidth]{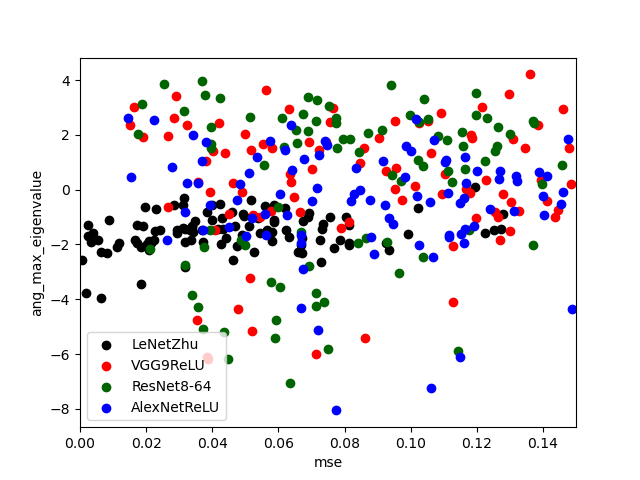}  
  \caption{ang\_max vs MSE (CS, $\sigma_{S} = 0.72$)}
  \label{fig:mse-angmax-cs}
\end{subfigure}
\begin{subfigure}{.325\textwidth}
  \centering
  \includegraphics[width=1\linewidth]{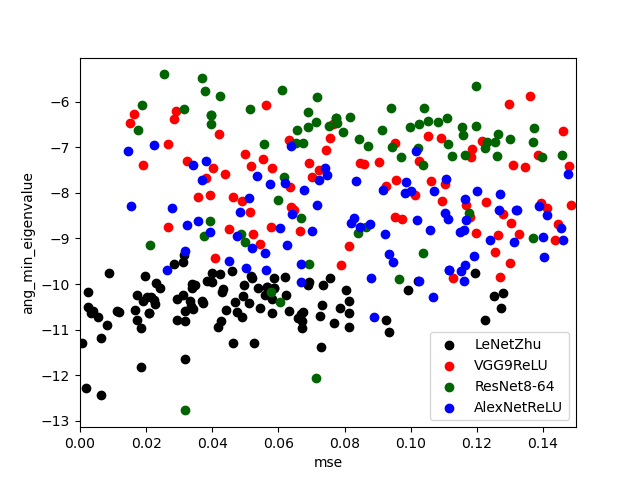}  
  \caption{ang\_min vs MSE (CS,  $\sigma_{S} = 0.75$))}
  \label{fig:corr-imagewoof}
\end{subfigure}

\caption{\textbf{Comparison of gradient norm, maximum and minimum eigenvalues of Hessian in terms of the correlation with MSE of reconstructed samples over several architectures on ImageWoof test samples.}}
  
\label{fig:imagewoof-corr}
\end{figure*}

We included plot results for the qualitative comparison between gradient norm and our proposed measures for datasets CIFAR100 (in Figure~\ref{fig:cifar100-corr}), ImageNette (in Figure~\ref{fig:imagenette-corr}), and ImageWoof (in Figure~\ref{fig:imagewoof-corr}).

\section{Understanding LAVP}
In this section, we investigate whether LAVP works as expected from our theory. To simulate the local optimization scenario, the initialized image in the attack is sampled from $x_{i} = x^{*} + 0.1sign(\mathcal{N})$, where $\mathcal{N}$ is the normal distribution and $sign$ is the sign function. This initialization scheme always ensures initial reconstruction error as MSE with $0.1$, thus simulating local optimization. In Table~\ref{tab:local-loss}, we present the final gradient matching loss average with its standard deviation in this setup. For each sample, we run the attack algorithm five times and SGD optimizer is used for optimization. Note that LAVP-L2 is aligned with $\mathcal{L}_{grad}^{final}$ (L2) and LAVP-cos is aligned with $\mathcal{L}_{grad}^{final}$ (cos) while gradient norm does not correlate with final loss for any kind of loss function.

\begin{table}[h]
\centering
\resizebox{0.9\columnwidth}{!}{
\centering
\begin{tabular}{|c|c|c|c|}

\hline
Values / Image index &     Image 1   &  Image 2 & Image 3 \\ \hline
$\mathcal{L}_{grad}^{final}$ (L2)  & 307.75 $\pm$ 15.45 & \textbf{0.001 $\pm$ 0.00} & \underline{526.42 $\pm$ 50.32} \\ \hline
$\mathcal{L}_{grad}^{final}$ (cos)  & \underline{0.053 $\pm$ 0.02} & 0.032 $\pm$ 0.00  &  \textbf{0.014 $\pm$ 0.00}\\ \hline
LAVP-L2(max) & 4.70E05  & \textbf{1.81} & \underline{2.19E06}\\ \hline
LAVP-L2(min) &  1.64 & \textbf{3.70E-06} & \underline{3.71}\\ \hline
LAVP-cos(ang$\_$max) & \underline{4.53}  & 0.99 & \textbf{0.08}\\ \hline
LAVP-cos(ang$\_$min) &  \underline{1.20E-03} & 4.00E-04 & \textbf{8.00E-05}\\ \hline
gradient norm &  24.64 & \textbf{0.07} & \underline{42.02}\\ \hline

\end{tabular}\par
}
\vspace{-6pt}
\caption{\textbf{Final gradient matching losses ($\mathcal{L}_{grad}^{final}$ (L2 distance), $\mathcal{L}_{grad}^{final}$ (cosine distance)), LAVPs for L2, LAVPs for cosine, and gradient norm for three different images. ) / LPIPS($\downarrow$)).} For each value, the smallest value among three images is marked as \textbf{bold} and the largest value among them is marked as \underline{underlined}. }
\label{tab:local-loss}
\end{table}
\end{document}